\documentclass[11pt]{article}

\usepackage[preprint]{acl}

\usepackage{times}
\usepackage{latexsym}

\usepackage[T1]{fontenc}

\usepackage[utf8]{inputenc}

\usepackage{microtype}

\usepackage{inconsolata}
\usepackage{tikz}
\usepackage{hyperref}
\usetikzlibrary{
  positioning,
  shapes.geometric,
  shadows,
  arrows.meta,
  backgrounds,
  calc
}

\newcommand{\CircledA}[1]{%
    \tikz[baseline=(char.base)]{
        \node[shape=circle,draw,inner sep=2pt, color=blue, fill=green] (char) {#1};}
}
\newcommand{\CircledC}[1]{%
    \tikz[baseline=(char.base)]{
        \node[shape=circle,draw,inner sep=2pt, color=blue, fill=red] (char) {#1};}
}

\usepackage{graphicx} 
\usepackage{booktabs} 
\usepackage{upquote} 
\usepackage{multirow} 
\usepackage{tabularx} 
\usepackage{arabtex} 
\usepackage{amsmath} 
\usepackage{amsfonts} 
\usepackage{amssymb} 
\usepackage{xcolor} 
\usepackage{colortbl} 
\usepackage{times} 
\usepackage{latexsym} 
\usepackage{enumitem} 
\usepackage{booktabs}
\usepackage{multirow} 
\usepackage{graphicx} 
\usepackage{amsfonts} 
\usepackage{amsmath} 
\usepackage{tikz} 
\usepackage{hyperref} 
\usepackage{tcolorbox}
\usepackage{bbm} 
\usepackage{xspace} 
\usepackage{subcaption}
\usepackage[colorinlistoftodos,prependcaption,textsize=tiny]{todonotes}
\usepackage{xspace}
\newcommand{\eg}{e.g.\@\xspace}
\usepackage{amsmath} 
\usepackage{textcomp} 
\usepackage{CJKutf8} 
\usepackage{arabtex} 
\usepackage{utf8} 
\usepackage{amssymb} 
\setcode{utf8} 
\newcommand{\dataset}{\textsc{MultiTempBench}\xspace}
\newcommand{\wz}[1]{\textcolor{black}{#1}}
\newcommand{\ga}[1]{\textcolor{black}{#1}}
\title{What Really Controls Temporal Reasoning in Large Language Models: \\ Tokenisation or  Representation of Time?}

\author{Gagan Bhatia\textsuperscript{1}\,
Ahmad Muhammad Isa\textsuperscript{1}\,
Maxime Peyrard\textsuperscript{2}\,
Wei Zhao\textsuperscript{1} \\[0.4em]
  \textsuperscript{1}University of Aberdeen \,
  \textsuperscript{2}Université Grenoble Alpes \& CNRS\\
  \texttt{wei.zhao@abdn.ac.uk}
  }

\begin{document}
\maketitle
\begin{abstract}
We present \dataset, a multilingual temporal reasoning benchmark spanning three tasks, date arithmetic, time zone conversion, and temporal relation extraction across five languages (English, German, Chinese, Arabic, and Hausa) and multiple calendar conventions (Gregorian, Hijri, and Chinese Lunar). \dataset contains $15{,}000$ examples built by translating $750$ curated English questions and expanding each into controlled date-format variants. We evaluate 20 LLMs and introduce the \emph{multilingual Date Fragmentation Ratio} (mDFR), calibrated with human severity ratings, together with \wz{geometric-probing analyses of internal temporal representations}.
We find tokenisation quality of temporal artefacts is a resource-dependent bottleneck: in low-resource languages and rarer calendar formats, fragmentation disrupts Year/Month/Day separation and accuracy collapses, while high-resource settings are often robust to digit-level splitting. Beyond tokenisation, crossed mixed-effects regression shows that temporal linearity is the strongest predictor of temporal reasoning in high-resource languages, whereas fragmentation is the stronger predictor in low-resource languages.
\footnote{https://github.com/gagan3012/mtb}

\end{list}
\end{abstract}

\section{Introduction}

Time is a universal substrate for human reasoning, but temporal expressions are deeply \emph{language-} and \emph{culture-}specific. Real-world systems—calendar assistants, travel planners, clinical and legal timeline reconstruction, historical question answering, and forecasting—must interpret and manipulate dates, times, and temporal relations expressed in heterogeneous surface forms (e.g., \texttt{2024-05-01} vs.\ ``1 May 2024'') and under distinct calendrical conventions (e.g., Gregorian vs.\ Hijri vs.\ Lunar). These requirements are inherently multilingual: users routinely mix scripts, localized month lexemes, and calendar markers, and many high-stakes workflows depend on correct temporal normalisation and arithmetic across languages and regions. Recent temporal benchmarks have advanced our understanding of LLMs’ abilities in date arithmetic, temporal ordering, and time-sensitive QA, but they overwhelmingly focus on English and Gregorian representations \citep{wang2024tram,fatemi2024tot,zhu2024freshbench,chu2023timebench,islakoglu2025chronosense,wei2025time,liu2025temporaltokenizationstrategiesevent,sasaki-etal-2025-language,pezik2025llmlagbenchidentifyingtemporaltraining}. In parallel, emerging work on \emph{cross-calendar} reasoning highlights that LLMs remain inadequate for inter-calendar conversion and that non-Gregorian temporal structure is still underexplored despite its global relevance \citep{han2025,saxena2025lostintime,miao2025span,wang2026measuringiterativetemporalreasoning,holtermann-etal-2025-around}.

\begin{figure}[t] 
\centering 
\includegraphics[width=1\linewidth]{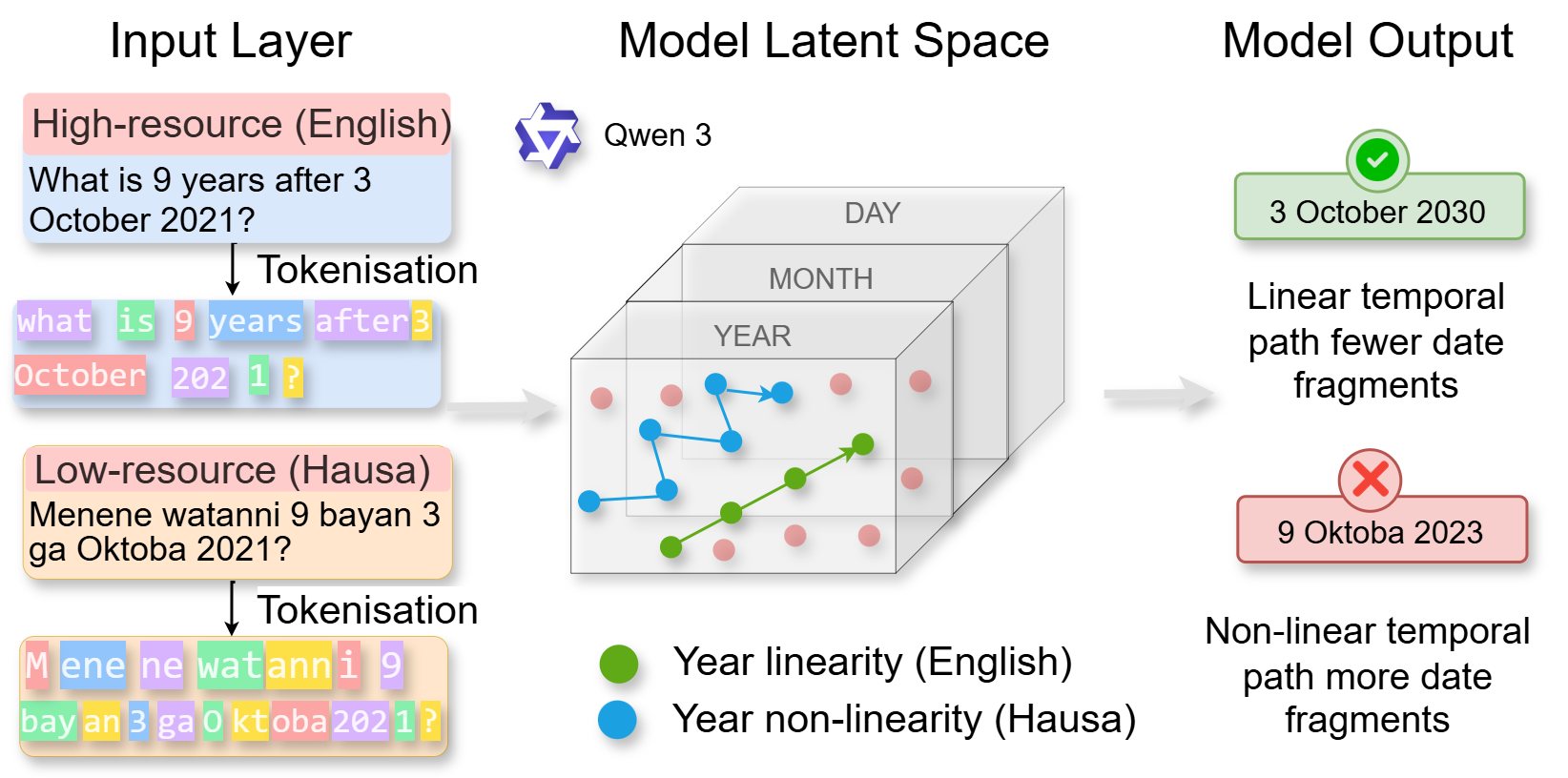}
\caption{\textbf{Mechanistic understanding of multilingual temporal reasoning in \dataset.} 
}
\label{fig:geometric_tax_overview} 
\end{figure}

An orthogonal challenge concerns \emph{how dates are presented to the model}. Temporal strings are structured symbolic objects, yet subword tokenisers (e.g., BPE and byte-level tokenisation) can fragment them into opaque substrings, potentially erasing semantic boundaries such as year/month/day separators and calendar markers \citep{spathis2023first,bhatia-etal-2025-date}. More broadly, multilingual tokenisation is known to induce systematic inequities: low-resource languages often incur heavier fragmentation (a ``token tax''), which increases effective sequence length and can degrade downstream performance \citep{ahia2023languages,petrov2023language,lundin2025tokentax,kanjirangat2025tokenization}. For numeracy specifically, the choice between digit-level tokenisation and larger numeric chunks yields distinct arithmetic failure modes: inconsistent segmentation forces the model to infer place value and grouping from unstable boundaries, complicating the learning of multi-digit operations and carry-like mechanisms \citep{singh2024tokenizationcounts,kreitner2025efficientnumeracy}.
These observations motivate a natural hypothesis: multilingual temporal failures may be primarily a \emph{tokenisation problem}. However, two gaps prevent a mechanistic account. First, existing tokenisation analyses typically target general text or broad downstream tasks, rather than \emph{controlled temporal expressions} that combine digits, delimiters, month lexemes, and calendar markers across scripts and calendar systems. Even temporally focused studies largely remain monolingual or calendar-specific \citep{bhatia-etal-2025-date,han2025,miao2025span}. Second, even when behavioural gaps are documented (e.g., accuracy differences and systematic error patterns across languages and formats), we lack clarity on \emph{where} the failure arises in the processing pipeline: does temporal information degrade at the input layer, fail to form an abstract representation suitable for computation, or break during reasoning and decoding? Prior mechanistic work suggests that LLMs can encode ordered scalar attributes in ways that are approximately linearly decodable from hidden states, and can exhibit stable latent directions corresponding to monotonic temporal progression \citep{gurnee2024spacetime,el-shangiti-etal-2025-geometry}, but these findings have not been connected to multilingual, multi-calendar temporal reasoning in a controlled setting. As a result, it remains unclear whether multilingual temporal competence requires (i) better surface segmentation, (ii) a shared internal ``calendar geometry'' that supports computation, or (iii) both.

Our contributions are summarized as follows:
\begin{enumerate}[label=(\roman*), leftmargin=*]
    \item \textbf{A controlled multilingual, multi-calendar benchmark.} We release \dataset, comprising 15{,}000 examples across 5 languages, 3 temporal tasks, and multiple date-format complexity levels, including Gregorian, Hijri, and Lunar calendar systems. We evaluate a broad set of LLMs (open-weight and proprietary) in a zero-shot setting, quantifying how language resource level, format complexity, and calendar system affect temporal reasoning.
    \item \textbf{A multilingual fragmentation metric.} We propose \textbf{mDFR}, a multilingual extension of Date Fragmentation Ratio \citep{bhatia-etal-2025-date} that penalises semantically destructive segmentations (e.g., digit splitting and boundary loss) with weights learned from human severity ratings.
    \item \textbf{A mechanistic account of multilingual temporal performance gaps.} We show that bottlenecks shift by language: \emph{date fragmentation} is most predictive of failure in low-resource regimes (where it disrupts access to compositional components), whereas \emph{temporal linearity} (probe $R^2$) is the strongest \wz{predictor of temporal task performance}     
    in high-resource languages once calendar 
    \wz{components (e.g., Year/Month/Day)} are \wz{frequently} accessible in \wz{training data}. This supports a two-stage view in which tokenisation 
    \wz{controls an LLM's surface-level access to calendar components while temporal linearity controls internal temporal representations.} To test this directly, we complement descriptive analyses with a crossed mixed-effects regression over all model predictions, allowing us to compare the contribution of date fragmentation and temporal linearity across resource levels.
\end{enumerate}

\section{Related Works}
\label{sec:related}

\paragraph{Tokenisation bias in multilingual models.}
Tokenisation remains a critical source of disparity in multilingual LLMs. Recent studies confirm that low-resource languages and dialects suffer a token tax, inflated sequence lengths that degrade performance and increase compute costs \citep{lundin2025tokentax, kanjirangat2025tokenization}. This disparity is particularly acute for Indian languages, where standard vocabularies often fragment morphological units \citep{karthika2025indian}. Such fragmentation \wz{may} critically impact numeric and temporal reasoning. \citet{bhatia-etal-2025-date} identify date fragmentation as a hidden bottleneck, showing that Byte-Pair Encoding (BPE) often splits dates into opaque substrings that hinder temporal arithmetic. Similarly, \citet{singh2024tokenizationcounts} demonstrate that standard tokenisation degrades arithmetic performance compared to single-token number embeddings. Our work extends this line of inquiry by introducing \dataset to systematically isolate how tokenisation \wz{quality of dates},
measured via our proposed mDFR metric, \wz{affect}
reasoning capabilities in a multilingual setting.

\paragraph{Mechanisms of time: memorisation vs. reasoning.}
\ga{Evaluating temporal understanding requires distinguishing between pattern matching and robust reasoning. Indeed, recent studies reveal that while LLMs maintain stable performance on memorisation-based temporal tasks, their accuracy sharply declines on reasoning-intensive tasks, especially when navigating temporal shifts or integrating new knowledge \citep{mazzia-etal-2026-benchmarking, li2025memorization, li2025diagnosing}. Benchmarks such as ChronoSense \citep{islakoglu2025chronosense}, SPAN \cite{miao2025spanbenchmarkingimprovingcrosscalendar}, DateLogicQA \citep{bhatia-etal-2025-datelogicqa} and TimeBench \citep{chu2023timebench} reveal that LLMs struggle with symbolic constraints and temporal commonsense. Mechanistically, \citet{gurnee2024spacetime} find that models possess linear subspaces representing space and time, suggesting relevant information is encoded but not always exploited. However, \citet{mamidanna2025allforone} observe that computation is often aggregated only at the final token, creating a fragile information bottleneck. We situate \dataset{} at the intersection of these fields. Unlike broad benchmarks, we use controlled date expressions to disentangle tokenisation from reasoning.}

\section{Our \dataset}
\label{sec:dataset}

\begin{table}[!ht]
\centering
\setlength{\tabcolsep}{3.5pt} 
\renewcommand{\arraystretch}{0.95} 
\resizebox{\columnwidth}{!}{ 
\footnotesize
\begin{tabular}{lllc}
\toprule
\textbf{Lang. (Size)} & \textbf{Type} & \textbf{Pattern} & \textbf{Example} \\
\midrule
\multirow{4}{*}{\shortstack[l]{English\\(300GB)}} 
 & ISO & YYYY-MM-DD & 2023-07-03 \\
 & Numeric & DD/MM/YYYY & 03/07/2023 \\
 & Textual & DD Month YYYY & 03 July 2023 \\
 & Phrasal & Day of Month YYYY & 3rd of July 2023 \\
\cmidrule{1-4}
\multirow{4}{*}{\shortstack[l]{German\\(66GB)}} 
 & ISO & YYYY-MM-DD & 2023-07-03 \\
 & Numeric & DD.MM.YYYY & 03.07.2023 \\
 & Textual & DD. Month YYYY & 03. Juli 2023 \\
 & Phrasal & DD. Mon... YYYY & 03. Juli des Jahres 2023 \\
\cmidrule{1-4}
\multirow{4}{*}{\shortstack[l]{Chinese\\(47GB)}} 
 & ISO & YYYY-MM-DD & 2023-07-03 \\
 & Numeric & DD/MM/YYYY & 03/07/2023 \\
 & Textual & \begin{CJK*}{UTF8}{gbsn}Y 年 M 月 D 日\end{CJK*} & \begin{CJK*}{UTF8}{gbsn}2023年07月03日\end{CJK*} \\
 & Lunar & \textit{Traditional} & \begin{CJK*}{UTF8}{gbsn}二零二三年六月初九\end{CJK*} \\
\cmidrule{1-4}
\multirow{4}{*}{\shortstack[l]{Arabic\\(28GB)}} 
 & ISO & YYYY-MM-DD & 2023-07-03 \\
 & Numeric & DD/MM/YYYY & 03/07/2023 \\
 & Textual & DD Month YYYY & \RL{٣ يوليو ٢٠٢٣} \\
 & Hijri & Hijri DD Mon YYYY & \RL{٣ ربيع الأول ٤٠٥هـ} \\
\cmidrule{1-4}
\multirow{4}{*}{\shortstack[l]{Hausa\\(0.3GB)}} 
 & ISO & YYYY-MM-DD & 2023-07-03 \\
 & Numeric & DD/MM/YYYY & 03/07/2023 \\
 & Textual & DD ga Month YYYY & 03 ga Yuli 2023 \\
 & Hijri & DD Mon YYYY AH & 03 Ramadan 1445 AH \\
\bottomrule
\end{tabular}
}
\caption{\textbf{Date formats and calendar systems in \dataset.} The ``Type'' column indicates the format category or specific calendar system (e.g., Lunar, Hijri). All others use the Gregorian calendar.}
\label{tab:merged-date-formats}
\end{table}

\paragraph{Dataset construction.} 
We introduce \dataset, a multilingual temporal reasoning benchmark derived from three existing datasets: TRAM \citep{wang2024tram}, ToT \citep{fatemi2024tot}, and FreshBench \citep{zhu2024freshbench}. TRAM contains 526,668 multiple-choice questions across 10 temporal reasoning tasks covering the period from 1000 to 2024. ToT consists of 46,480 questions focusing on temporal semantics and arithmetic from 52 AD to 2087. FreshBench provides 4,643 forecasting questions from 1900 to 2025. To construct the English foundation of \dataset, we curated a balanced subset of 750 questions: 250 from TRAM, 250 from ToT, and 250 from FreshBench, \wz{covering three temporal reasoning tasks: (i) \textbf{Date Arithmetic}, which evaluates the ability of LLMs to perform addition and subtraction on dates; (ii) \textbf{Time Zone Conversion}, which tests the understanding of LLMs to calculate time differences between regions 
and (iii) \textbf{Temporal Relation}, which infers the relationship (e.g., before, after, simultaneous) between a specific event and a reference date. Data samples in these tasks are provided in Table \ref{tab:tasks} (appendix).} We selected questions where date components (year, month, and day) are fully specified, then we preprocessed them to remove synthetic entities (e.g., ``E15'') and internal prompting instructions, ensuring all questions are grammatically correct and natural.

\paragraph{Multilingual extension.} 
We extended these 750 English questions into four additional languages: German, Chinese, Hausa, and Arabic, \wz{using the Google Translate \cite{comanici2025gemini25pushingfrontier}. 
We manually verified the machine-generated translations, for each target language, two native speakers were involved to validate the translations and edited them (when necessary) to ensure that both the linguistic content and date formats were error-free}. The set of languages we selected is based on our linguistic expertise, as well as diversified data availability on the CommonCrawl-100 corpus \cite{suarez2019asynchronous,penedo2025fineweb2}, ranging from
high-resource languages like English (300 GB) to low-resource ones like Hausa (0.3 GB). As detailed in Table \ref{tab:merged-date-formats}, these languages also cover three calendar systems: Gregorian, Hijri, and Chinese Lunar for 
temporal reasoning.

\paragraph{Data format extension.} 
To assess robustness \wz{of temporal reasoning} across date formats, we utilized a template-based approach to expand each question into four variants per language with increasing levels of complexity. As shown in Table \ref{tab:merged-date-formats}, these formats range from \textit{standard ISO Numeric} (e.g., YYYY-MM-DD) to \textit{Localised Numeric} formats using local separators, and finally to \textit{Calendar-specific} phrases (e.g., ``03 Ramadan 1445 AH'' for Hausa). For calendar-specific variants (e.g., Hijri and Chinese Lunar), we 
converted to target-language calendars 
by using 
existing calender conversion tools \cite{alshehri_2024_11288565}; the results were verified by native speakers.
This 
expansion results 
in 3,000 questions per language, totalling 15,000 questions.
For a detailed description of the conversion tools, library specifications, and language-specific formatting rules, we refer to Appendix \ref{app:date_pipeline}.

\section{Our Approach}
\label{sec:method}
\wz{Our aim is to identify the underlying factors that control temporal reasoning, then evaluate how these factors vary across languages. To do so, we present a metric, which we call the multilingual Date Fragmentation Ratio (mDFR), to measure the tokenisation quality of dates, then introduce temporal geometry to capture the geometric structures of internal temporal representations.}
\subsection{Multilingual Date Fragmentation Ratio}
We extend the Date Fragmentation Ratio (DFR) from \citet{bhatia-etal-2025-date} to a multilingual setup, which we call mDFR denoted as $F \in [0,1]$: 
\begin{equation}
F = \alpha_1 \mathbbm{1}_{\mathrm{split}} + \alpha_2 \mathbbm{1}_{\mathrm{delimiter}} + \alpha_3 \Delta N + \alpha_4 \theta
\end{equation}

Here, $\mathbbm{1}_{\mathrm{split}}$ and $\mathbbm{1}_{\mathrm{delimiter}}$ are binary indicators for split semantic roots (e.g., splitting ``2024'') and lost separators, respectively. $\Delta N$ represents the token count inflation relative to a semantic baseline. Finally, $\theta$ quantifies the structural divergence between the model's token distribution and the ideal semantic units using cosine distance, as defined in \citet{bhatia-etal-2025-date}. We calibrate the coefficients $\alpha$ by fitting a linear model to human judgements of fragmentation severity across our target languages. We perform human evaluation of mDFR in Appendix \ref{app:metric_validation}.

\subsection{Temporal Geometry}
\label{ssec:temporal_geometry_method}
\paragraph{Embedding extraction.}
For each language $\ell \in \{\textsc{en},\textsc{de},\textsc{zh},\textsc{ar},\textsc{ha}\}$ and year $y \in [1990,2024]$ \ga{in 3 different date formats (ISO, Slash and Long)}
we aim to extract a robust representation of the year that is invariant to specific months or days. To do so, we sample $K=5$ distinct full dates within year $y$ (e.g., ``1995-03-12'', ``1995-11-05'') and embed them into declarative templates (e.g., ``The date is \textit{<date>}'' or \mbox{\RL{التاريخ هو}} \textit{<date>}). We propagate these sequences through the model and extract the hidden state $\mathbf{h}^{(\ell)}_{y,k,i} \in \mathbb{R}^d$ corresponding to the final token at layer $i$ for the $k$-th date sample of year $y$. We define the average embedding as:
\begin{equation}
\bar{\mathbf{h}}^{(\ell)}_{y,i} = \frac{1}{K} \sum_{k=1}^{K} \mathbf{h}^{(\ell)}_{y,k,i}.
\end{equation}

\paragraph{Geometric notations.}
We use several geometric concepts to describe the geometry of time in the embedding space.
\begin{itemize}
    \item \textbf{Line segment.} $\mathbf{s}^{(\ell)}_{y,i}$ is a line segment connecting two vectors $\bar{\mathbf{h}}^{(\ell)}_{y+1,i}$ and $\bar{\mathbf{h}}^{(\ell)}_{y,i}$, indicating the vector difference between two years $y{+}1$ and $y$ in the embedding space:
    \begin{equation}
    \mathbf{s}^{(\ell)}_{y,i}
    =
    \bar{\mathbf{h}}^{(\ell)}_{y+1,i}-\bar{\mathbf{h}}^{(\ell)}_{y,i}.
    \label{eq:local_step_app_geom}
    \end{equation}

    \item \textbf{A path of line segments.} A sequence of line segments is denoted as
    \[
    \mathcal{P}^{(\ell)}_{i}
    =
    \big(\mathbf{s}^{(\ell)}_{y_1,i},\mathbf{s}^{(\ell)}_{y_2,i},\dots,\mathbf{s}^{(\ell)}_{y_T,i}\big)
    \]
    where each line segment connects to the next, forming a path of years from $1$ to $T$ in the embedding space.

    \item \textbf{The path direction.} We denote the overall path direction as the average of line segments:
    \begin{equation}
    \boldsymbol{\Delta}^{(\ell)}_{i}
    =
    \frac{1}{|Y|-1}\sum_{y}\mathbf{s}^{(\ell)}_{y,i}.
    \label{eq:mean_step_app_geom}
    \end{equation}
    If most line segments point in the same direction, then $\boldsymbol{\Delta}^{(\ell)}_{i}$ is stable and represents a clear ``forward-in-time'' direction for language $\ell$ at the $i$-th layer.
\end{itemize}



\paragraph{Linear structure of time.}
We test whether calendar values \wz{(e.g., a sequence of years \{2000, $\ldots$, 2010\})} form an underlying linear structure in a 1D subspace of the embedding space. For year calendar component $c \in \{Y\}$ (Year), we train a linear regressor that decodes the corresponding scalar value from the hidden representation $\bar{\mathbf{h}}^{(\ell)}_{y,i}$. Concretely, we fit
\begin{equation}
\hat{c} = \mathbf{W}_{c}\,\bar{\mathbf{h}}^{(\ell)}_{y,i} + \mathbf{b}_{c},
\qquad
\text{Linearity}(c) = R^2(c,\hat{c}),
\label{eqn:Linearity_general}
\end{equation}
where $R^2$ measures how well the \wz{Year} values can be recovered by a single linear readout. A higher $R^2$ indicates that the \wz{Year component} is organised along an approximately ordered axis in the embedding space, \wz{which may help LLMs perform date arithmetic more effectively. We also apply this idea to Month and Day components.}

\section{Experiments}
We examine a diverse suite of decoder-only LLMs to disentangle the effects of 
\wz{model size}, architecture, and tokeniser composition. 
Our open LLMs include the Qwen3 family (spanning from 0.6B to 14B parameters; \cite{yang2025qwen3technicalreport}), LLaMA~3 (8B, 70B; \cite{touvron2023llama}), and variants of OLMo \cite{olmo20252olmo2furious,groeneveld2024olmoacceleratingsciencelanguage}, Gemma \cite{gemmateam2025gemma3technicalreport}, Mistral \cite{mistralai2025magistral}, and Phi-4 \cite{microsoft2025phi4minitechnicalreportcompact}. To benchmark against proprietary 
systems, we evaluate \texttt{GPT-4o} and \texttt{GPT-4o-mini} \citep{openai2024gpt4ocard}. This selection allows us to isolate tokeniser-induced errors \citep{singh2024tokenizationcounts, lundin2025tokentaxsystematicbias,bhatia-etal-2025-date,bhatia-etal-2025-datelogicqa} from \wz{obscure} failures in 
temporal reasoning.

\subsection{Tokenisation Setup}
\label{ssec:token-method}

\paragraph{Baseline vs.\ model tokenisers.}
We contrast each model's native subword segmentation against a deterministic, linguistically informed \textit{baseline tokeniser}. This baseline segments date strings into semantic primitives (year, month, day, calendar marker), strictly preserving delimiters and whitespace. It is designed to be language-aware, correctly parsing Arabic-Indic numerals, Chinese temporal markers (e.g., \begin{CJK*}{UTF8}{gbsn}年, 月\end{CJK*}), and Hijri suffixes. For each instance in \dataset, we compute the divergence between the model's native segmentation (using TikToken or Hugging Face tokenizers) and this semantic baseline.

\begin{table*}[!htp]
\centering
\resizebox{\textwidth}{!}{
\begin{tabular}{llclllc}
\toprule
\textbf{Format} & \textbf{Language} & \textbf{Calendar} & \textbf{Original String} & \textbf{Baseline Tokenization} & \textbf{Gemma 3 Tokenization (Visualized)} & \textbf{mDFR} \\
\midrule
DD. Month YYYY & German & Greg. & 10. Oktober 2034 & 10 . Oktober 2034 & \texttt{1} $|$ \texttt{0} $|$ \texttt{.} $|$ \texttt{Oktober} $|$ \texttt{2} $|$ \texttt{0} $|$ \texttt{3} $|$ \texttt{4} & 0.50 \\
Month DD, YYYY & English & Greg. & October 10, 2034 & October 10 , 2034 & \texttt{October} $|$ \texttt{1} $|$ \texttt{0} $|$ \texttt{,} $|$ \texttt{2} $|$ \texttt{0} $|$ \texttt{3} $|$ \texttt{4} & 0.53 \\
\begin{CJK*}{UTF8}{gbsn}YYYY年MM月DD日\end{CJK*} & Chinese & Greg. & \begin{CJK*}{UTF8}{gbsn}2034 年 10 月 10 日\end{CJK*} & \begin{CJK*}{UTF8}{gbsn}2034 年 10 月 10 日\end{CJK*} & \texttt{2} $|$ \texttt{0} $|$ \texttt{3} $|$ \texttt{4} $|$ \begin{CJK*}{UTF8}{gbsn}年\end{CJK*} $|$ \texttt{1} $|$ \texttt{0} $|$ \begin{CJK*}{UTF8}{gbsn}月\end{CJK*} $|$ \texttt{1} $|$ \texttt{0} $|$ \begin{CJK*}{UTF8}{gbsn}日\end{CJK*} & 0.55 \\
DD Month YYYY \RL{هـ} & Arabic & Hijri & \RL{٢٧ رجب ١٤٥٦ هـ} & \RL{٢٧ رجب ١٤٥٦ هـ} & \RL{٢} $|$ \RL{٧} $|$ \RL{ر} $|$ \RL{جب} $|$ \RL{١} $|$ \RL{٤} $|$ \RL{٥} $|$ \RL{٦} $|$ \RL{هـ} & 0.60 \\
DD Month YYYY AH & English & Hijri & 27 Rajab 1456 AH & 27 Rajab 1456 AH & \texttt{2} $|$ \texttt{7} $|$ \texttt{Raj} $|$ \texttt{ab} $|$ \texttt{1} $|$ \texttt{4} $|$ \texttt{5} $|$ \texttt{6} $|$ \texttt{AH} & 0.60 \\
\begin{CJK*}{UTF8}{gbsn}干支年\end{CJK*} MM\begin{CJK*}{UTF8}{gbsn}月DD\end{CJK*} & Chinese & Lunar & \begin{CJK*}{UTF8}{gbsn}辛亥年 五月廿三\end{CJK*} & \begin{CJK*}{UTF8}{gbsn}辛亥年 五月廿三\end{CJK*} & \begin{CJK*}{UTF8}{gbsn}辛\end{CJK*}$|$ \begin{CJK*}{UTF8}{gbsn}亥\end{CJK*} $|$ \begin{CJK*}{UTF8}{gbsn}年\end{CJK*} $|$ \begin{CJK*}{UTF8}{gbsn}五\end{CJK*} $|$ \begin{CJK*}{UTF8}{gbsn}月\end{CJK*} $|$ \begin{CJK*}{UTF8}{gbsn}廿\end{CJK*} $|$ \begin{CJK*}{UTF8}{gbsn}三\end{CJK*} & 0.65 \\
DD Month YYYY & Arabic & Greg. & \RL{١٠ أكتوبر ٢٠٣٤} & \RL{١٠ أكتوبر ٢٠٣٤} & \RL{١} $|$ \RL{٠} $|$ \RL{أكتوبر} $|$ \RL{٢} $|$ \RL{٠} $|$ \RL{٣} $|$ \RL{٤} & 0.70 \\
DD Month YYYY & English & Greg. & 10 October 2034 & 10 October 2034 & \texttt{1} $|$ \texttt{0} $|$ \texttt{October} $|$ \texttt{2} $|$ \texttt{0} $|$ \texttt{3} $|$ \texttt{4} & 0.75 \\
Month DD, YYYY & Hausa & Greg. & Oktoba 10, 2034 & Oktoba 10 , 2034 & \texttt{O} $|$ \texttt{kt} $|$ \texttt{oba} $|$ \texttt{1} $|$ \texttt{0} $|$ \texttt{,} $|$ \texttt{2} $|$ \texttt{0} $|$ \texttt{3} $|$ \texttt{4} & 0.78 \\
\bottomrule
\end{tabular}
}
\caption{\textbf{Qualitative Analysis of Tokenisation Fragmentation.} Vertical bars ($|$) denote token boundaries within the Gemma 3 tokenizer. Note the severe fragmentation in non-Latin scripts (Arabic, Chinese) and the splitting of month names in Hausa.}
\label{tab:token_qualitative}
\end{table*}

\paragraph{Multilingual date fragmentation ratio (mDFR).}
We evaluate \wz{tokenisation quality}
using our \textbf{mDFR} metric. 
The learned coefficients for the metric are $\alpha = (0.2, 0.2, 0.1, 0.5)$, (Table \ref{tab:learned_weights_appendix}) reflecting that structural divergence ($\theta$), and root splitting are more detrimental than simple token count inflation. 
Table \ref{tab:token_qualitative} provides a qualitative comparison of tokenisation behaviours using the Gemma 3 tokeniser. 
High-resource languages like German and English exhibit moderate fragmentation (mDFR $\approx$ 0.50--0.53), typically characterised by the splitting of numeric roots (e.g., 2034'' becoming \texttt{2}$|$\texttt{0}$|$\texttt{3}$|$\texttt{4}) while largely preserving semantic delimiters and month names. In contrast, low-resource settings suffer from semantic fragmentation; for instance, the Hausa date Oktoba 10, 2034'' yields the highest DFR of 0.78, as the month name is broken into opaque sub-word units (\texttt{O}$|$\texttt{kt}$|$\texttt{oba}) alongside the numeric splitting.

\subsection{Temporal Reasoning Evaluation Setup}
\label{ssec:eval-method}

\paragraph{Prompting strategy.}
We evaluate models in a zero-shot setting without \wz{fine-tuning, chain-of-thought demonstrations, or external knowledge, as these may help LLMs resolve temporal tasks even if date strings are poorly tokenised.}
Each prompt consists of the question and a concise instruction to output the final answer.


\paragraph{LLM-as-a-judge.}
Given the \wz{diverse}
output formats across languages, we employ an LLM-based evaluation pipeline. For every prediction, we generate a JSON record containing the question, the model's raw output, and a set of gold-standard aliases (e.g., ``03/04/2025'', ``3 April 2025'', \mbox{\RL{٣ أبريل ٢٠٢٥}}). \texttt{GPT-4o} acts as the judge, classifying the response as \texttt{CORRECT} (consistent with gold aliases), \texttt{INCORRECT} (mutually exclusive), or \texttt{NOT\_ATTEMPTED}, \wz{which was initially introduced by OpenAI for the QA task \cite{wei2024measuring}}. The automated judge achieved a 87\% agreement rate with the majority human vote (Inter annotator agreement Cohen’s $\kappa = 0.89$) on a validation set of 250 multilingual instances. (For more details, please see Appendix~\ref{app:human_eval})


\section{Results}
\label{sec:results}
Our goal is to address the core
question: \emph{what controls temporal reasoning performance, surface tokenisation of dates, or internal geometric structures of temporal representations}? To do so,
we first test whether \textbf{date fragmentation} predicts accuracy across models and settings (Section~\ref{ssec:res-fragments}). We then test whether \textbf{calendar geometry} in hidden states predicts accuracy (Section~\ref{ssec:res-geometry}). Finally, we synthesise which factor is necessary and/or sufficient for strong temporal reasoning (Section~\ref{ssec:res-synthesis}).

\subsection{Multilingual Temporal Reasoning Performance}
\label{ssec:res-accuracy}

Table~\ref{tab:language-model-results} reports temporal reasoning accuracy averaged across the three tasks within each language. Two patterns stand out. First, performance is highly language-dependent: most models are relatively strong in high-resource languages (English, Chinese, German) but degrade sharply in Hausa, indicating a distinct low-resource regime where temporal reasoning is brittle.
\ga{Second, model ranking is not explained by model size alone: some smaller open-source LLMs outperform larger ones (for example, the 4B-parameter Gemma 3 achieves a 59.2\% average, surpassing both the 8B-parameter Llama 3.1 at 57.3\% and the 20B-parameter GPT-OSS at 20.0\%), suggesting that multilingual coverage and training/tokenisation choices outrank raw parameter count for this benchmark.}
These accuracy trends motivate the mechanistic split we test next: low-resource failures align with an \emph{input accessibility} bottleneck (date fragmentation), whereas high-resource variation is better explained by an \emph{internal geometry} bottleneck (temporal linearity).

\begin{figure*}[t!]
    \centering
    \includegraphics[width=\textwidth]{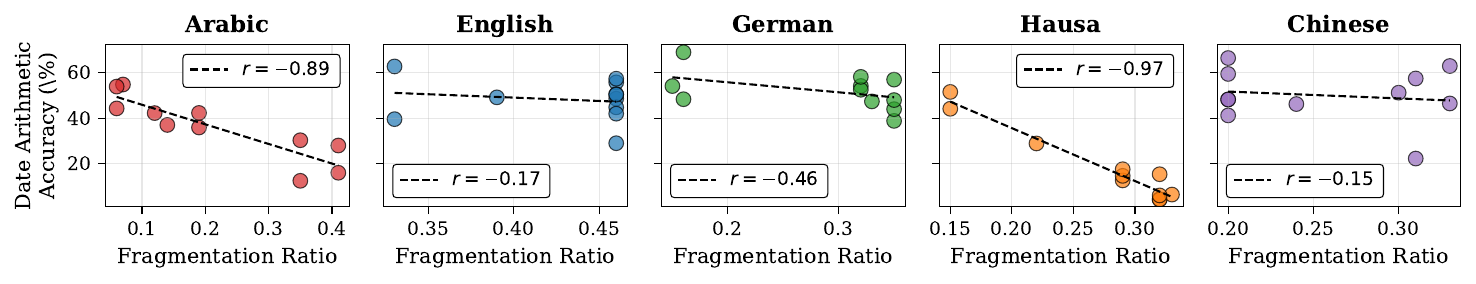}
    \caption{\textbf{Impact of Tokenisation on Date Arithmetic Accuracy.} DFR is strongly negatively correlated with accuracy in Hausa ($r=-0.97$), but only weakly correlated in English ($r=-0.17$).
    }
    \label{fig:dfr_correlation}
\end{figure*}

\begin{table}[!t]
\centering
\resizebox{\columnwidth}{!}{
\begin{tabular}{lrrrrrr}
\toprule
& \multicolumn{5}{c}{\textbf{Accuracy by Language (\%)}} & \\
\cmidrule(lr){2-6}
\textbf{Model} & \textbf{Arabic} & \textbf{Chinese} & \textbf{English} & \textbf{German} & \textbf{Hausa} & \textbf{Average} \\
\midrule
\rowcolor{gray!10} \multicolumn{7}{l}{\textit{Proprietary Models}} \\
GPT-4o & \textbf{71.3} & \textbf{66.0} & 54.3 & \textbf{70.0} & \textbf{51.7} & \textbf{62.7} \\
\midrule
\rowcolor{gray!10} \multicolumn{7}{l}{\textit{Open-Weights Models}} \\
Gemma 3 4B & 57.3 & 64.7 & 63.7 & 64.0 & 46.3 & \textbf{59.2} \\
Llama 3.1 8B & 49.0 & 65.0 & \textbf{66.7} & 64.3 & 41.3 & 57.3 \\
Phi-4 Mini & 39.7 & 55.7 & 66.3 & 62.7 & 28.7 & 50.6 \\
Qwen 3 4B & 41.3 & 56.7 & 54.7 & 46.0 & 9.0 & 41.5 \\
Mistral 7B v0.2 & 44.3 & 40.7 & 51.7 & 54.7 & 9.0 & 40.1 \\
Llama 2 7B & 15.7 & 40.7 & 55.0 & 51.3 & 17.0 & 35.9 \\
Gemma 3 1B & 27.3 & 42.7 & 40.7 & 38.0 & 19.3 & 33.6 \\
Olmo 3 7B	&16.3&	39.3&	48.0&	33.3&	12.3&	29.8 \\
DS-R1 Qwen 7B & 24.7 & 48.0 & 45.3 & 42.3 & 1.7 & 32.4 \\
OLMo 2 7B & 16.3 & 39.3 & 48.0 & 33.3 & 12.3 & 29.9 \\
GPT-OSS 20B&	5.0&	24.0&	49.0&	20.3&	2.0&	20.0 \\
Qwen 3 14B & 25.3 & 9.7 & 19.0 & 27.7 & 2.3 & 16.8 \\
Qwen3 0.6B	&21.7&	19.7 &	14.7&	23.3&	4.0&	16.6 \\
\bottomrule
\end{tabular}}
\caption{\textbf{Multilingual Temporal Reasoning Accuracy.} Accuracy is averaged across the three tasks (date arithmetic, time zone conversion, temporal relation extraction) within each language. }
\label{tab:language-model-results}
\end{table}

\begin{table}[!htp]
\centering
\resizebox{\columnwidth}{!}{
\begin{tabular}{lrrrrrrrrr}
\toprule
\textbf{Calendar} & \multicolumn{5}{c}{\textbf{Gregorian}} & \multicolumn{1}{c}{\textbf{Lunar}} & \multicolumn{3}{c}{\textbf{Hijri}} \\
\cmidrule(lr){2-6} \cmidrule(lr){7-7} \cmidrule(lr){8-10}
\textbf{Model} & \textbf{Ar} & \textbf{Zh} & \textbf{En} & \textbf{De} & \textbf{Ha} & \textbf{Zh} & \textbf{Ar} & \textbf{En} & \textbf{Ha} \\
\midrule
Baseline     & 0.00 & 0.00 & 0.00 & 0.00 & 0.00 & 0.00 & 0.00 & 0.00 & 0.00 \\
GPT-3.5      & 0.19 & 0.12 & 0.23 & 0.12 & 0.12 & 0.41 & 0.14 & 0.31 & 0.30 \\
GPT-4        & 0.19 & 0.12 & 0.23 & 0.12 & 0.12 & 0.41 & 0.14 & 0.31 & 0.30 \\
GPT-5        & 0.19 & 0.12 & 0.23 & 0.12 & 0.12 & 0.41 & 0.14 & 0.31 & 0.30 \\
Llama 2      & 0.06 & 0.23 & 0.42 & 0.30 & 0.29 & 0.20 & 0.04 & 0.23 & 0.23 \\
Phi 3.5      & 0.06 & 0.23 & 0.42 & 0.30 & 0.29 & 0.20 & 0.04 & 0.23 & 0.23 \\
Mistral      & 0.06 & 0.23 & 0.42 & 0.30 & 0.29 & 0.29 & 0.04 & 0.23 & 0.23 \\
Davinci-003  & 0.17 & 0.17 & 0.37 & 0.10 & 0.16 & 0.29 & 0.09 & 0.37 & 0.39 \\
OLMo         & 0.13 & 0.18 & 0.37 & 0.09 & 0.16 & 0.41 & 0.09 & 0.37 & 0.40 \\
Llama 3      & 0.35 & 0.16 & 0.34 & 0.12 & 0.12 & 0.60 & 0.36 & 0.31 & 0.30 \\
DeepSeek     & 0.10 & 0.31 & 0.44 & 0.34 & 0.32 & 0.61 & 0.05 & 0.30 & 0.29 \\
gpt-oss      & 0.39 & 0.16 & 0.34 & 0.12 & 0.13 & 0.60 & 0.44 & 0.32 & 0.32 \\
Qwen3        & 0.17 & 0.32 & 0.44 & 0.34 & 0.32 & 0.61 & 0.16 & 0.31 & 0.28 \\
Cohere       & 0.18 & 0.34 & 0.44 & 0.34 & 0.33 & 0.63 & 0.22 & 0.32 & 0.32 \\
Gemma3       & 0.39 & 0.34 & 0.44 & 0.34 & 0.33 & 0.48 & 0.30 & 0.31 & 0.31 \\
\bottomrule
\end{tabular}
}
\caption{Multilingual Date Fragmentation Ratio (mDFR) across models for Gregorian (Ar, Zh, En, De, Ha), Lunar (Zh), and Hijri (Ar, En, Ha). Higher scores indicate greater fragmentation of date tokens.}
\label{tab:dfr_results}
\end{table}

\subsection{Date Fragments and Temporal Reasoning}
\label{ssec:res-fragments}

\paragraph{Fragmentation varies by language and calendar format.}
Table~\ref{tab:dfr_results} reports mDFR across models and calendar varieties. Fragmentation arises from digit splitting (\eg, \texttt{2034} $\rightarrow$ \texttt{2|0|3|4})(Table~\ref{tab:token_qualitative}). These effects are amplified in low-resource settings and in less frequent calendar variants: for instance,
non-Gregorian formats are often incorrectly tokenised, with \wz{more} date fragments than Gregorian formats (Table~\ref{tab:token_qualitative}), due to sparse calendar markers and lower-frequency month lexemes.


\paragraph{Date fragments are a major bottleneck for temporal reasoning in low-resource languages.}
Figure~\ref{fig:dfr_correlation} shows that the greater the mDFR, the lower the accuracy in the date arithmetic task for two low-resource languages: Hausa ($r=-0.97$) and Arabic ($r=-0.89$). However, the correlation becomes much weaker for the three high-resource languages: German, Chinese, and English, indicating that more date fragments do not cause an accuracy collapse for these languages. Overall, these results suggest that tokenisation is a major bottleneck for temporal reasoning in low-resource languages, but not in high-resource ones:
\citet{bhatia-etal-2025-date} explained why this is the case in English: they found that LLMs can compensate for date fragmentation by stitching fragmented date tokens during temporal reasoning. We speculate that in high-resource languages, such fragments are still frequently observed in training data, which enables LLMs, especially larger models, to internally address/stitch them. \ga{We observe similar findings in the other two tasks: time zone conversion and temporal relation (see Figure~\ref{fig:dfr_correlation_tr} and Figure~\ref{fig:dfr_correlation_tz}).}

\subsection{Geometric Structures and Temporal Reasoning}
\label{ssec:res-geometry}

Tokenisation characterises what information is presented to the model, but not whether the model organises that information in a form suitable for computation. We therefore test whether \wz{temporal representations possess} 
internal geometric structures that support calendar manipulation. Here we focus on \textbf{temporal linearity}: how well temporal values (e.g., years) are organised along approximately ordered 1D axes, measured by the $R^2$ of linear probes decoding \emph{Day}, \emph{Month}, and \emph{Year} from hidden states.

\begin{figure*}[t]
    \centering
    \includegraphics[width=\textwidth]{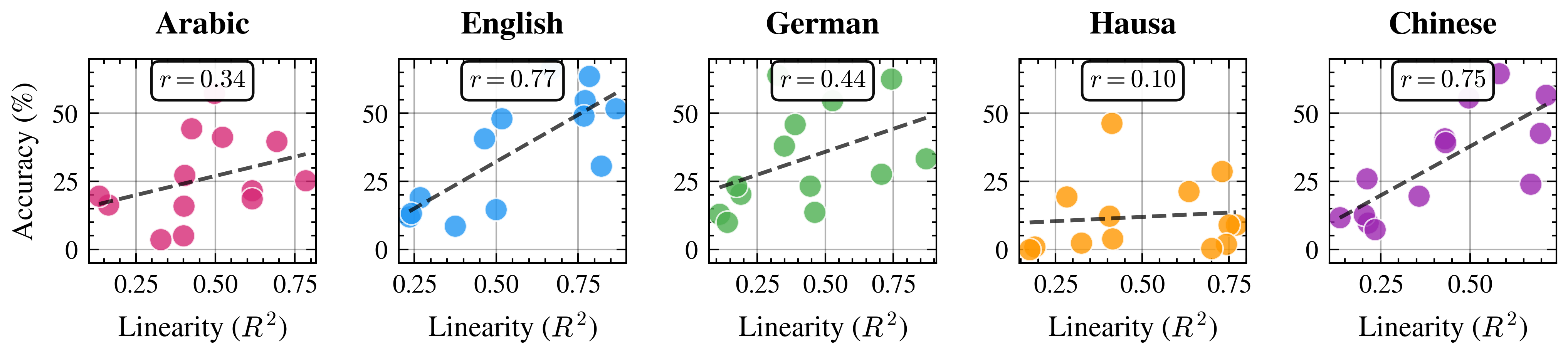}
    \caption{\textbf{Temporal linearity vs.\ accuracy across languages.} Temporal linearity (probe $R^2$) is strongly correlated with accuracy in English ($r{=}0.77$) and Chinese ($r{=}0.75$), but weakly correlated in Hausa ($r{=}0.10$), suggesting that ordered temporal geometry is a key driver of high performance when it emerges.}
    \label{fig:linearity_accuracy}
\end{figure*}

\paragraph{Temporal linearity is a strong \wz{predictor} of performance in high-resource languages.}
\ga{Figure~\ref{fig:linearity_accuracy} plots the average temporal reasoning accuracy (across date arithmetic, time zone conversion, and temporal relation extraction) against overall temporal linearity (aggregated across all calendar components, including delimiters) across models within each language. Overall, temporal linearity is strongly correlated with accuracy in English ($r=0.77$) and Chinese ($r=0.75$), moderately correlated in German ($r=0.44$) and Arabic ($r=0.34$), and weakly correlated in Hausa ($r=0.10$). This pattern suggests that, once models can reliably access calendar components, strong performance depends on representing temporal \emph{values} in an ordered geometry that supports arithmetic-like updates. In other words, in high-resource settings the main limiter is not surface form, but whether the model embeds time on a usable internal axis.}


\begin{figure}[t]
    \centering
    \includegraphics[width=\linewidth]{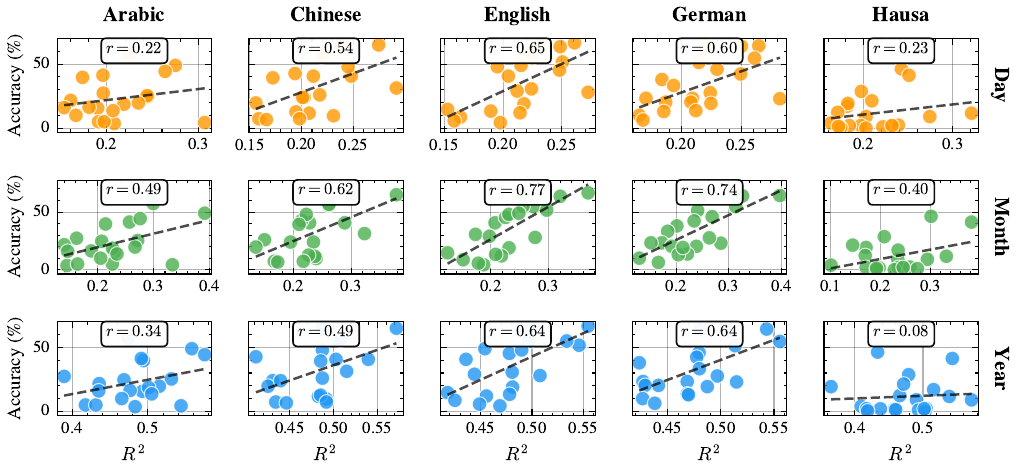}
    \caption{
    \textbf{Component-wise temporal linearity vs.\ accuracy.} Correlations between accuracy and probe $R^2$ for \textbf{Day}, \textbf{Month}, and \textbf{Year} within each language. 
    }
\label{fig:component_linearity_accuracy}
\end{figure}

\paragraph{Component-wise view: the Year axis is typically the most predictive.}
\ga{While Figure~\ref{fig:linearity_accuracy} summarises the holistic relationship between temporal linearity and accuracy, noting that this overall correlation is distinct from a simple average of the individual components, Figure~\ref{fig:component_linearity_accuracy} decomposes this geometry by calendar component (Day/Month/Year) within each language. Two trends stand out. First, correlations are generally strongest for \textbf{Year} (especially in English/Chinese), consistent with year values providing the primary ordered backbone required for many temporal operations. Second, \textbf{Month} and \textbf{Day} linearity show weaker and more heterogeneous correlations across languages. This suggests that Month and Day representations are not as robustly formed as Year representations, and are instead more sensitive to language- and format-specific cues (e.g., month lexemes and delimiters) than to a universal ordered axis. Overall, the component-wise breakdown supports the interpretation that ordered temporal geometry matters most when it provides a stable \emph{year} backbone, and that this signal is clearest in high-resource languages.}


\subsection{Which Mechanism Controls Temporal Reasoning?}
\label{ssec:res-synthesis}

\begin{figure}[t]
    \centering
    \includegraphics[width=\linewidth]{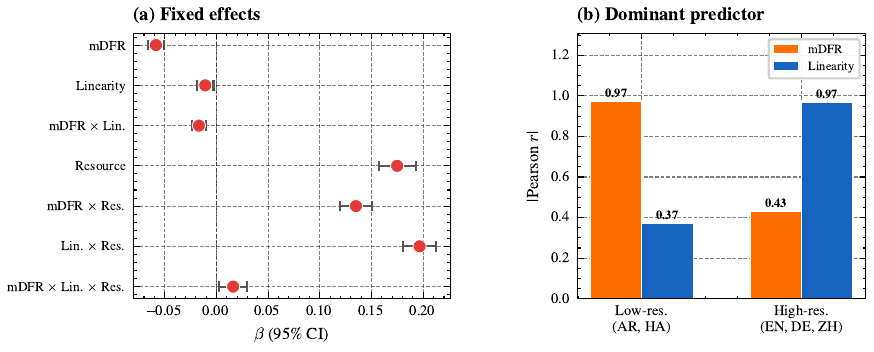}
    \caption{\textbf{Mixed-effects summary of temporal reasoning bottlenecks.} (a) Fixed effects from the crossed mixed-effects regression. (b) Dominant predictor by resource regime: mDFR in low-resource languages, linearity in high-resource languages.}
    \label{fig:bottleneck_summary}
\end{figure}

\ga{
To test the relative contribution of \textbf{date fragmentation} and \textbf{temporal linearity}, we fit a crossed mixed-effects regression predicting \textbf{per-question accuracy} over all 285000 predictions (15000 questions $\times$ 19 models). The dependent variable was binary correctness for each prediction. As fixed effects, we included z-scored \textbf{mDFR}, z-scored \textbf{linearity}, \textbf{resource level} (high-resource vs.\ low-resource), and all interaction terms:
$\texttt{correct} \sim \texttt{mDFR}_z * \texttt{linearity}_z * \texttt{resource}.$
We also included crossed random intercepts for \textbf{question} and \textbf{model} to account for item difficulty and model-specific baseline performance. This analysis lets us test whether temporal reasoning performance is better explained by \textbf{surface tokenisation of dates} or by \textbf{internal geometric structures of temporal representations}, and whether this differs by resource level.
The regression confirms that temporal reasoning performance is governed by a \textbf{language-dependent bottleneck}. We report regression coefficients using the notation $\beta$ (coefficient), $SE$ (standard error), $z$ (Wald statistic), and $p$ ($p$-value). Most importantly, the three-way interaction between \textbf{mDFR}, \textbf{linearity}, and \textbf{resource level} is significant ($\beta=0.016$, $SE=0.007$, $z=2.31$, $p=0.021$), showing that the dominant predictor changes across language regimes. In \textbf{low-resource} languages (Arabic and Hausa), higher fragmentation strongly predicts lower accuracy ($\beta=-0.126$, $p<0.001$), indicating that \textbf{date fragmentation} is the dominant bottleneck. In \textbf{high-resource} languages (English, German, and Chinese), \textbf{temporal linearity} is instead the stronger predictor of accuracy ($\beta=0.087$, $p<0.001$), while mDFR has only a weak effect ($\beta=0.009$, $p=0.056$). These results align with the language-wise analyses above: \textbf{low-resource languages tend to be input-limited}, whereas \textbf{high-resource languages tend to be geometry-limited}.  \textcolor{black}{Figure~\ref{fig:bottleneck_summary}a summarises these results across the \dataset. }
Figure~\ref{fig:bottleneck_summary}a also shows that the main effects of resource level and its interactions with mDFR and linearity are the largest fixed effects in the model, consistent with a resource-dependent shift in the dominant bottleneck. As shown in Figure~\ref{fig:bottleneck_summary}b, this yields a clear split: \textbf{low-resource languages tend to be input-limited}, whereas \textbf{high-resource languages tend to be geometry-limited}.
Overall, no single factor universally controls temporal reasoning across languages; instead, the dominant constraint shifts from \textbf{date fragmentation} to \textbf{temporal linearity} as resource level increases.
}
\textcolor{black}{Practically, this distinction matters because resource gaps across languages are expensive and slow to close, whereas linearity gaps may be more tractable through targeted interventions, such as re-aligning temporal representations.}

\section{Conclusion}
\label{sec:conclusion}

\dataset shows that multilingual temporal intelligence depends on more than adding vocabulary: it requires making temporal information \emph{accessible} and \emph{computable} in the model’s internal space. A crossed mixed-effects regression confirms this language-dependent bottleneck: in low-resource regimes, date fragmentation is the stronger predictor of failure, while in high-resource regimes temporal linearity is the stronger predictor of performance.


\section*{Limitations}
\textsc{MultiTempBench} is designed as a controlled diagnostic, and that design imposes constraints on generality: it covers five languages (English, German, Chinese, Arabic, and Hausa) and three task families (date arithmetic, time zone conversion, and temporal relation extraction), so it does not fully represent the diversity of multilingual temporal phenomena (e.g., additional scripts and dialects, code-mixing/noisy text, domain-specific jargon, or other calendar conventions beyond those included); instances are produced via translation and templated format variation from a curated English seed set, which helps isolate surface-form/tokenisation effects but may under-sample naturally occurring distributions of expressions and errors; we evaluate in a zero-shot, direct-answer setting (and normalise outputs with an LLM-as-a-judge), which improves comparability yet may understate performance under tool use, prompting, or fine-tuning and introduces residual evaluation noise from judge mistakes and format ambiguity; and while we find strong associations between fragmentation metrics, temporal linearity, and performance, our mechanistic analyses remain correlational and probe-centric, leaving open causal questions about tokeniser design, training data, decoding dynamics, and non-linear representational structure. A key limitation of our multilingual setup is that the low-resource regime is represented by only two languages (Arabic and Hausa), so claims about low-resource temporal reasoning should be interpreted as suggestive rather than fully general across low-resource languages. More broadly, the language split into high-resource versus low-resource is necessarily coarse, and additional languages are needed to test whether the same bottleneck pattern holds across other typological profiles, scripts, and calendar traditions.

\section*{Ethical Considerations}
This benchmark surfaces disparities that can arise from multilingual tokenisation and resource imbalance, but such results should be framed as properties of model design and data coverage rather than as inherent deficits of particular languages to avoid reinforcing harmful narratives; because calendar expressions are culturally situated (including non-Gregorian systems such as Hijri and Chinese Lunar), conversion or formatting errors can have real consequences in downstream, potentially high-stakes contexts, so users should document conversion assumptions and validate systems with native-speaker and domain-expert review when decisions matter; the dataset is constructed from public sources through translation and controlled transformations and is not intended to contain personal data, yet extensions should avoid introducing identifiable or sensitive information and, the human annotation is used follows informed consent and fair compensation practices; finally, any dependence on third-party model APIs for translation and/or evaluation can affect reproducibility and raise governance concerns, so releases will document versions and settings and, where feasible, provide open alternatives, while acknowledging that improved temporal reasoning can be dual-use and warrants domain-specific risk assessment and human oversight in sensitive deployments.

\section*{Broader Impact}
By providing a controlled multilingual temporal benchmark and analysis signals (e.g., fragmentation and representation geometry probes), this work can help the community audit and improve temporal reasoning across scripts, languages, and calendar conventions, potentially reducing ``token tax'' effects and improving language equity in multilingual NLP; it may also guide more principled tokeniser and training-data interventions by linking surface segmentation properties to downstream competence; however, like any benchmark, it can distort incentives if treated as a leaderboard target, encouraging optimisation for templated formats or discouraging support for languages that score poorly, so we emphasise its role as a diagnostic instrument rather than a deployment-readiness test and encourage follow-on work to broaden coverage (more languages/dialects and naturalistic temporal text), evaluate mitigation strategies directly, and report results with uncertainty and careful error analysis to support responsible, inclusive progress.

\bibliography{custom}

@article{wei2024measuring,
  title={Measuring short-form factuality in large language models},
  author={Wei, Jason and Karina, Nguyen and Chung, Hyung Won and Jiao, Yunxin Joy and Papay, Spencer and Glaese, Amelia and Schulman, John and Fedus, William},
  journal={arXiv preprint arXiv:2411.04368},
  year={2024}
}

@inproceedings{kanjirangat2025tokenization,
  author    = {Vani Kanjirangat and Tanja Samard{\v{z}}i{\'c} and Ljiljana Dolamic and Fabio Rinaldi},
  title     = {Tokenization and Representation Biases in Multilingual Models on Dialectal {NLP} Tasks},
  booktitle = {Proceedings of the 2025 Conference on Empirical Methods in Natural Language Processing},
  year      = {2025},
  note      = {EMNLP 2025},
  url       = {https://arxiv.org/abs/2509.20045}
}

@article{singh2024tokenizationcounts,
  author  = {Aaditya K. Singh and D. J. Strouse},
  title   = {Tokenization Counts: The Impact of Tokenization on Arithmetic in Frontier {LLMs}},
  journal = {arXiv preprint arXiv:2402.14903},
  year    = {2024},
  url     = {https://arxiv.org/abs/2402.14903}
}

@article{lundin2025tokentax,
  author  = {Jessica M. Lundin and Ada Zhang and Nihal Karim and Hamza Louzan and Victor Wei and David Adelani and Cody Carroll},
  title   = {The Token Tax: Systematic Bias in Multilingual Tokenization},
  journal = {arXiv preprint arXiv:2509.05486},
  year    = {2025},
  url     = {https://arxiv.org/abs/2509.05486}
}

@article{kreitner2025efficientnumeracy,
  author  = {Linus Kreitner and Paul Hager and Jonathan Mengedoht and Georgios Kaissis and Daniel Rueckert and Martin J. Menten},
  title   = {Efficient Numeracy in Language Models through Single-Token Number Embeddings},
  journal = {arXiv preprint arXiv:2510.06824},
  year    = {2025},
  url     = {https://arxiv.org/abs/2510.06824}
}

@article{karthika2025indian,
  author  = {N. J. Karthika and Maharaj Brahma and Rohit Saluja and Ganesh Ramakrishnan and Maunendra Sankar Desarkar},
  title   = {Multilingual Tokenization through the Lens of Indian Languages: Challenges and Insights},
  journal = {arXiv preprint arXiv:2506.17789},
  year    = {2025},
  url     = {https://arxiv.org/abs/2506.17789}
}

@inproceedings{islakoglu2025chronosense,
  author    = {Duygu Sezen Islakoglu and Jan{-}Christoph Kalo},
  title     = {ChronoSense: Exploring Temporal Understanding in Large Language Models with Time Intervals of Events},
  booktitle = {Proceedings of the 63rd Annual Meeting of the Association for Computational Linguistics (Short Papers)},
  year      = {2025},
  note      = {ACL 2025},
  url       = {https://arxiv.org/abs/2501.03040}
}

@inproceedings{li2025diagnosing,
  author    = {Huihan Li and You Chen and Siyuan Wang and Yixin He and Ninareh Mehrabi and Rahul Gupta and Xiang Ren},
  title     = {Diagnosing Memorization in Chain-of-Thought Reasoning, One Token at a Time},
  booktitle = {Proceedings of the 2025 Conference on Empirical Methods in Natural Language Processing},
  year      = {2025},
  note      = {EMNLP 2025},
  url       = {https://arxiv.org/abs/2508.02037}
}

@inproceedings{mamidanna2025allforone,
  author    = {Siddarth Mamidanna and Daking Rai and Ziyu Yao and Yilun Zhou},
  title     = {All for One: {LLMs} Solve Mental Math at the Last Token With Information Transferred From Other Tokens},
  booktitle = {Proceedings of the 2025 Conference on Empirical Methods in Natural Language Processing},
  year      = {2025},
  note      = {EMNLP 2025},
  url       = {https://arxiv.org/abs/2509.09650}
}

@inproceedings{gurnee2024spacetime,
  author    = {Wes Gurnee and Max Tegmark},
  title     = {Language Models Represent Space and Time},
  booktitle = {Proceedings of the 12th International Conference on Learning Representations},
  year      = {2024},
  note      = {ICLR 2024},
  url       = {https://arxiv.org/abs/2310.02207}
}

@inproceedings{li2025memorization,
  author    = {Aochong Oliver Li and Tanya Goyal},
  title     = {Memorization vs. Reasoning: Updating {LLMs} with New Knowledge},
  booktitle = {Findings of the Association for Computational Linguistics: ACL 2025},
  year      = {2025},
  note      = {ACL Findings 2025},
  url       = {https://arxiv.org/abs/2504.12523}
}

@article{chu2023timebench,
  author  = {Zheng Chu and Jingchang Chen and Qianglong Chen and Weijiang Yu and Haotian Wang and Ming Liu and Bing Qin},
  title   = {TimeBench: A Comprehensive Evaluation of Temporal Reasoning Abilities in Large Language Models},
  journal = {arXiv preprint arXiv:2311.17667},
  year    = {2023},
  url     = {https://arxiv.org/abs/2311.17667}
}

@article{saxena2025lostintime,
  author  = {Rohit Saxena and Aryo Pradipta Gema and Pasquale Minervini},
  title   = {Lost in Time: Clock and Calendar Understanding Challenges in Multimodal {LLMs}},
  journal = {arXiv preprint arXiv:2502.05092},
  year    = {2025},
  url     = {https://arxiv.org/abs/2502.05092}
}

@inproceedings{wei2025time,
  author    = {Shaohang Wei and Wei Li and Feifan Song and Wen Luo and Tianyi Zhuang and Haochen Tan and Zhijiang Guo and Houfeng Wang},
  title     = {TIME: A Multi-level Benchmark for Temporal Reasoning of {LLMs} in Real-World Scenarios},
  booktitle = {Advances in Neural Information Processing Systems},
  year      = {2025},
  note      = {NeurIPS 2025},
  url       = {https://arxiv.org/abs/2505.12891}
}

@article{han2025,
  author    = {Yicheng Han and Shih-Ming Wang and Jialu Zhang and Qian Liu and Wei Lu},
  title     = {Ticktack: Modeling temporal relationships in LLMs using non-Gregorian calendars},
  journal   = {arXiv preprint arXiv:2503.04150},
  year      = {2025},
  url       = {https://arxiv.org/abs/2503.04150}
}

@article{spathis2023first,
  title={The first step is the hardest: Pitfalls of Representing and Tokenizing Temporal Data for Large Language Models},
  author={Spathis, Dimitris and Kawsar, Fahim},
  journal={arXiv preprint arXiv:2309.06236},
  year={2023},
  url={https://arxiv.org/abs/2309.06236}
}

@article{petrov2023language,
  title={Language Model Tokenizers Introduce Unfairness Between Languages},
  author={Petrov, Aleksandar and La Malfa, Emanuele and Torr, Philip HS and Bibi, Adel},
  journal={arXiv preprint arXiv:2305.15425},
  year={2023},
  eprint={2305.15425},
  archivePrefix={arXiv},
  primaryClass={cs.CL},
  url={https://arxiv.org/abs/2305.15425}
}

@article{ahia2023languages,
  title={Do All Languages Cost the Same? Tokenization in the Era of Commercial Language Models},
  author={Ahia, Orevaoghene and de Almeida, David and Shleifer, Nathan and Dinan, Emily},
  journal={arXiv preprint arXiv:2305.13707},
  year={2023},
  eprint={2305.13707},
  archivePrefix={arXiv},
  primaryClass={cs.CL},
  url={https://arxiv.org/abs/2305.13707}
}

@inproceedings{wang2024tram,
  title={TRAM: Benchmarking Temporal Reasoning for Large Language Models},
  author={Wang, Yuqing and Zhao, Yun},
  booktitle={Findings of the Association for Computational Linguistics: ACL 2024},
  year={2024},
  url={https://arxiv.org/abs/2310.00835}
}

@article{fatemi2024tot,
  title={Test of Time: A Benchmark for Evaluating LLMs on Temporal Reasoning},
  author={Fatemi, Bahare and Kazemi, Mehran and Tsitsulin, Anton and Malkan, Karishma and Yim, Jinyeong and Palowitch, John and Seo, Sungyong and Halcrow, Jonathan and Perozzi, Bryan},
  journal={arXiv preprint arXiv:2406.09170},
  year={2024},
  url={https://arxiv.org/abs/2406.09170}
}

@inproceedings{zhu2024freshbench,
  title={{Evaluating LLMs at Evaluating Temporal Generalization}},
  author={Zhu, Chenghao and Chen, Nuo and Gao, Yufei and Zhang, Yunyi and Tiwari, Prayag and Wang, Benyou},
  booktitle={Proceedings of the 2025 Conference of the North American Chapter of the Association for Computational Linguistics: Human Language Technologies (Volume 1: Long Papers)},
  year={2024},
  url={https://arxiv.org/html/2405.08460v1}
}

@article{touvron2023llama,
  title={LLaMA: Open and Efficient Foundation Language Models},
  author={Touvron, Hugo and Lavril, Thibaut and Izacard, Gautier and Martinet, Xavier and Lachaux, Marie-Anne and Lacroix, Timothée and Rozière, Baptiste and Goyal, Naman and Hambro, Eric and Azhar, Faisal and others},
  journal={arXiv preprint},
  year={2023},
  doi={10.48550/arXiv.2302.13971},
  url={https://arxiv.org/abs/2302.13971}
}

@misc{miao2025span,
  title         = {Benchmarking and Improving Cross-Calendar Temporal Reasoning of Large Language Models},
  author        = {Miao, Zeyu and others},
  year          = {2025},
  eprint        = {2511.09993},
  archivePrefix = {arXiv},
  primaryClass  = {cs.CL},
  url           = {https://arxiv.org/abs/2511.09993}
}

@inproceedings{el-shangiti-etal-2025-geometry,
  title     = {The Geometry of Numerical Reasoning: Language Models Compare Numeric Properties in Linear Subspaces},
  author    = {El-Shangiti, Ahmed Oumar and Hiraoka, Tatsuya and AlQuabeh, Hilal and Heinzerling, Benjamin and Inui, Kentaro},
  booktitle = {Proceedings of the 2025 Conference of the Nations of the Americas Chapter of the Association for Computational Linguistics: Human Language Technologies (Volume 2: Short Papers)},
  month     = apr,
  year      = {2025},
  address   = {Albuquerque, New Mexico},
  publisher = {Association for Computational Linguistics},
  url       = {https://aclanthology.org/2025.naacl-short.47/},
  doi       = {10.18653/v1/2025.naacl-short.47},
  pages     = {550--561}
}

@misc{yang2025qwen3technicalreport,
      title={Qwen3 Technical Report}, 
      author={An Yang and Anfeng Li and Baosong Yang and Beichen Zhang and Binyuan Hui and Bo Zheng and Bowen Yu and Chang Gao and Chengen Huang and Chenxu Lv and Chujie Zheng and Dayiheng Liu and Fan Zhou and Fei Huang and Feng Hu and Hao Ge and Haoran Wei and Huan Lin and Jialong Tang and Jian Yang and Jianhong Tu and Jianwei Zhang and Jianxin Yang and Jiaxi Yang and Jing Zhou and Jingren Zhou and Junyang Lin and Kai Dang and Keqin Bao and Kexin Yang and Le Yu and Lianghao Deng and Mei Li and Mingfeng Xue and Mingze Li and Pei Zhang and Peng Wang and Qin Zhu and Rui Men and Ruize Gao and Shixuan Liu and Shuang Luo and Tianhao Li and Tianyi Tang and Wenbiao Yin and Xingzhang Ren and Xinyu Wang and Xinyu Zhang and Xuancheng Ren and Yang Fan and Yang Su and Yichang Zhang and Yinger Zhang and Yu Wan and Yuqiong Liu and Zekun Wang and Zeyu Cui and Zhenru Zhang and Zhipeng Zhou and Zihan Qiu},
      year={2025},
      eprint={2505.09388},
      archivePrefix={arXiv},
      primaryClass={cs.CL},
      url={https://arxiv.org/abs/2505.09388}, 
}

@misc{openai2024gpt4ocard,
      title={GPT-4o System Card}, 
      author={OpenAI and : and Aaron Hurst and Adam Lerer and Adam P. Goucher and Adam Perelman and Aditya Ramesh and Aidan Clark and AJ Ostrow and Akila Welihinda and Alan Hayes and Alec Radford and Aleksander Mądry and Alex Baker-Whitcomb and Alex Beutel and Alex Borzunov and Alex Carney and Alex Chow and Alex Kirillov and Alex Nichol and Alex Paino and Alex Renzin and Alex Tachard Passos and Alexander Kirillov and Alexi Christakis and Alexis Conneau and Ali Kamali and Allan Jabri and Allison Moyer and Allison Tam and Amadou Crookes and Amin Tootoochian and Amin Tootoonchian and Ananya Kumar and Andrea Vallone and Andrej Karpathy and Andrew Braunstein and Andrew Cann and Andrew Codispoti and Andrew Galu and Andrew Kondrich and Andrew Tulloch and Andrey Mishchenko and Angela Baek and Angela Jiang and Antoine Pelisse and Antonia Woodford and Anuj Gosalia and Arka Dhar and Ashley Pantuliano and Avi Nayak and Avital Oliver and Barret Zoph and Behrooz Ghorbani and Ben Leimberger and Ben Rossen and Ben Sokolowsky and Ben Wang and Benjamin Zweig and Beth Hoover and Blake Samic and Bob McGrew and Bobby Spero and Bogo Giertler and Bowen Cheng and Brad Lightcap and Brandon Walkin and Brendan Quinn and Brian Guarraci and Brian Hsu and Bright Kellogg and Brydon Eastman and Camillo Lugaresi and Carroll Wainwright and Cary Bassin and Cary Hudson and Casey Chu and Chad Nelson and Chak Li and Chan Jun Shern and Channing Conger and Charlotte Barette and Chelsea Voss and Chen Ding and Cheng Lu and Chong Zhang and Chris Beaumont and Chris Hallacy and Chris Koch and Christian Gibson and Christina Kim and Christine Choi and Christine McLeavey and Christopher Hesse and Claudia Fischer and Clemens Winter and Coley Czarnecki and Colin Jarvis and Colin Wei and Constantin Koumouzelis and Dane Sherburn and Daniel Kappler and Daniel Levin and Daniel Levy and David Carr and David Farhi and David Mely and David Robinson and David Sasaki and Denny Jin and Dev Valladares and Dimitris Tsipras and Doug Li and Duc Phong Nguyen and Duncan Findlay and Edede Oiwoh and Edmund Wong and Ehsan Asdar and Elizabeth Proehl and Elizabeth Yang and Eric Antonow and Eric Kramer and Eric Peterson and Eric Sigler and Eric Wallace and Eugene Brevdo and Evan Mays and Farzad Khorasani and Felipe Petroski Such and Filippo Raso and Francis Zhang and Fred von Lohmann and Freddie Sulit and Gabriel Goh and Gene Oden and Geoff Salmon and Giulio Starace and Greg Brockman and Hadi Salman and Haiming Bao and Haitang Hu and Hannah Wong and Haoyu Wang and Heather Schmidt and Heather Whitney and Heewoo Jun and Hendrik Kirchner and Henrique Ponde de Oliveira Pinto and Hongyu Ren and Huiwen Chang and Hyung Won Chung and Ian Kivlichan and Ian O'Connell and Ian O'Connell and Ian Osband and Ian Silber and Ian Sohl and Ibrahim Okuyucu and Ikai Lan and Ilya Kostrikov and Ilya Sutskever and Ingmar Kanitscheider and Ishaan Gulrajani and Jacob Coxon and Jacob Menick and Jakub Pachocki and James Aung and James Betker and James Crooks and James Lennon and Jamie Kiros and Jan Leike and Jane Park and Jason Kwon and Jason Phang and Jason Teplitz and Jason Wei and Jason Wolfe and Jay Chen and Jeff Harris and Jenia Varavva and Jessica Gan Lee and Jessica Shieh and Ji Lin and Jiahui Yu and Jiayi Weng and Jie Tang and Jieqi Yu and Joanne Jang and Joaquin Quinonero Candela and Joe Beutler and Joe Landers and Joel Parish and Johannes Heidecke and John Schulman and Jonathan Lachman and Jonathan McKay and Jonathan Uesato and Jonathan Ward and Jong Wook Kim and Joost Huizinga and Jordan Sitkin and Jos Kraaijeveld and Josh Gross and Josh Kaplan and Josh Snyder and Joshua Achiam and Joy Jiao and Joyce Lee and Juntang Zhuang and Justyn Harriman and Kai Fricke and Kai Hayashi and Karan Singhal and Katy Shi and Kavin Karthik and Kayla Wood and Kendra Rimbach and Kenny Hsu and Kenny Nguyen and Keren Gu-Lemberg and Kevin Button and Kevin Liu and Kiel Howe and Krithika Muthukumar and Kyle Luther and Lama Ahmad and Larry Kai and Lauren Itow and Lauren Workman and Leher Pathak and Leo Chen and Li Jing and Lia Guy and Liam Fedus and Liang Zhou and Lien Mamitsuka and Lilian Weng and Lindsay McCallum and Lindsey Held and Long Ouyang and Louis Feuvrier and Lu Zhang and Lukas Kondraciuk and Lukasz Kaiser and Luke Hewitt and Luke Metz and Lyric Doshi and Mada Aflak and Maddie Simens and Madelaine Boyd and Madeleine Thompson and Marat Dukhan and Mark Chen and Mark Gray and Mark Hudnall and Marvin Zhang and Marwan Aljubeh and Mateusz Litwin and Matthew Zeng and Max Johnson and Maya Shetty and Mayank Gupta and Meghan Shah and Mehmet Yatbaz and Meng Jia Yang and Mengchao Zhong and Mia Glaese and Mianna Chen and Michael Janner and Michael Lampe and Michael Petrov and Michael Wu and Michele Wang and Michelle Fradin and Michelle Pokrass and Miguel Castro and Miguel Oom Temudo de Castro and Mikhail Pavlov and Miles Brundage and Miles Wang and Minal Khan and Mira Murati and Mo Bavarian and Molly Lin and Murat Yesildal and Nacho Soto and Natalia Gimelshein and Natalie Cone and Natalie Staudacher and Natalie Summers and Natan LaFontaine and Neil Chowdhury and Nick Ryder and Nick Stathas and Nick Turley and Nik Tezak and Niko Felix and Nithanth Kudige and Nitish Keskar and Noah Deutsch and Noel Bundick and Nora Puckett and Ofir Nachum and Ola Okelola and Oleg Boiko and Oleg Murk and Oliver Jaffe and Olivia Watkins and Olivier Godement and Owen Campbell-Moore and Patrick Chao and Paul McMillan and Pavel Belov and Peng Su and Peter Bak and Peter Bakkum and Peter Deng and Peter Dolan and Peter Hoeschele and Peter Welinder and Phil Tillet and Philip Pronin and Philippe Tillet and Prafulla Dhariwal and Qiming Yuan and Rachel Dias and Rachel Lim and Rahul Arora and Rajan Troll and Randall Lin and Rapha Gontijo Lopes and Raul Puri and Reah Miyara and Reimar Leike and Renaud Gaubert and Reza Zamani and Ricky Wang and Rob Donnelly and Rob Honsby and Rocky Smith and Rohan Sahai and Rohit Ramchandani and Romain Huet and Rory Carmichael and Rowan Zellers and Roy Chen and Ruby Chen and Ruslan Nigmatullin and Ryan Cheu and Saachi Jain and Sam Altman and Sam Schoenholz and Sam Toizer and Samuel Miserendino and Sandhini Agarwal and Sara Culver and Scott Ethersmith and Scott Gray and Sean Grove and Sean Metzger and Shamez Hermani and Shantanu Jain and Shengjia Zhao and Sherwin Wu and Shino Jomoto and Shirong Wu and Shuaiqi and Xia and Sonia Phene and Spencer Papay and Srinivas Narayanan and Steve Coffey and Steve Lee and Stewart Hall and Suchir Balaji and Tal Broda and Tal Stramer and Tao Xu and Tarun Gogineni and Taya Christianson and Ted Sanders and Tejal Patwardhan and Thomas Cunninghman and Thomas Degry and Thomas Dimson and Thomas Raoux and Thomas Shadwell and Tianhao Zheng and Todd Underwood and Todor Markov and Toki Sherbakov and Tom Rubin and Tom Stasi and Tomer Kaftan and Tristan Heywood and Troy Peterson and Tyce Walters and Tyna Eloundou and Valerie Qi and Veit Moeller and Vinnie Monaco and Vishal Kuo and Vlad Fomenko and Wayne Chang and Weiyi Zheng and Wenda Zhou and Wesam Manassra and Will Sheu and Wojciech Zaremba and Yash Patil and Yilei Qian and Yongjik Kim and Youlong Cheng and Yu Zhang and Yuchen He and Yuchen Zhang and Yujia Jin and Yunxing Dai and Yury Malkov},
      year={2024},
      eprint={2410.21276},
      archivePrefix={arXiv},
      primaryClass={cs.CL},
      url={https://arxiv.org/abs/2410.21276}, 
}

@misc{lundin2025tokentaxsystematicbias,
      title={The Token Tax: Systematic Bias in Multilingual Tokenization}, 
      author={Jessica M. Lundin and Ada Zhang and Nihal Karim and Hamza Louzan and Victor Wei and David Adelani and Cody Carroll},
      year={2025},
      eprint={2509.05486},
      archivePrefix={arXiv},
      primaryClass={cs.CL},
      url={https://arxiv.org/abs/2509.05486}, 
}

@misc{groeneveld2024olmoacceleratingsciencelanguage,
      title={OLMo: Accelerating the Science of Language Models}, 
      author={Dirk Groeneveld and Iz Beltagy and Pete Walsh and Akshita Bhagia and Rodney Kinney and Oyvind Tafjord and Ananya Harsh Jha and Hamish Ivison and Ian Magnusson and Yizhong Wang and Shane Arora and David Atkinson and Russell Authur and Khyathi Raghavi Chandu and Arman Cohan and Jennifer Dumas and Yanai Elazar and Yuling Gu and Jack Hessel and Tushar Khot and William Merrill and Jacob Morrison and Niklas Muennighoff and Aakanksha Naik and Crystal Nam and Matthew E. Peters and Valentina Pyatkin and Abhilasha Ravichander and Dustin Schwenk and Saurabh Shah and Will Smith and Emma Strubell and Nishant Subramani and Mitchell Wortsman and Pradeep Dasigi and Nathan Lambert and Kyle Richardson and Luke Zettlemoyer and Jesse Dodge and Kyle Lo and Luca Soldaini and Noah A. Smith and Hannaneh Hajishirzi},
      year={2024},
      eprint={2402.00838},
      archivePrefix={arXiv},
      primaryClass={cs.CL},
      url={https://arxiv.org/abs/2402.00838}, 
}

@misc{olmo20252olmo2furious,
      title={2 OLMo 2 Furious}, 
      author={Team OLMo and Pete Walsh and Luca Soldaini and Dirk Groeneveld and Kyle Lo and Shane Arora and Akshita Bhagia and Yuling Gu and Shengyi Huang and Matt Jordan and Nathan Lambert and Dustin Schwenk and Oyvind Tafjord and Taira Anderson and David Atkinson and Faeze Brahman and Christopher Clark and Pradeep Dasigi and Nouha Dziri and Allyson Ettinger and Michal Guerquin and David Heineman and Hamish Ivison and Pang Wei Koh and Jiacheng Liu and Saumya Malik and William Merrill and Lester James V. Miranda and Jacob Morrison and Tyler Murray and Crystal Nam and Jake Poznanski and Valentina Pyatkin and Aman Rangapur and Michael Schmitz and Sam Skjonsberg and David Wadden and Christopher Wilhelm and Michael Wilson and Luke Zettlemoyer and Ali Farhadi and Noah A. Smith and Hannaneh Hajishirzi},
      year={2025},
      eprint={2501.00656},
      archivePrefix={arXiv},
      primaryClass={cs.CL},
      url={https://arxiv.org/abs/2501.00656}, 
}

@misc{gemmateam2025gemma3technicalreport,
      title={Gemma 3 Technical Report}, 
      author={Gemma Team and Aishwarya Kamath and Johan Ferret and Shreya Pathak and Nino Vieillard and Ramona Merhej and Sarah Perrin and Tatiana Matejovicova and Alexandre Ramé and Morgane Rivière and Louis Rouillard and Thomas Mesnard and Geoffrey Cideron and Jean-bastien Grill and Sabela Ramos and Edouard Yvinec and Michelle Casbon and Etienne Pot and Ivo Penchev and Gaël Liu and Francesco Visin and Kathleen Kenealy and Lucas Beyer and Xiaohai Zhai and Anton Tsitsulin and Robert Busa-Fekete and Alex Feng and Noveen Sachdeva and Benjamin Coleman and Yi Gao and Basil Mustafa and Iain Barr and Emilio Parisotto and David Tian and Matan Eyal and Colin Cherry and Jan-Thorsten Peter and Danila Sinopalnikov and Surya Bhupatiraju and Rishabh Agarwal and Mehran Kazemi and Dan Malkin and Ravin Kumar and David Vilar and Idan Brusilovsky and Jiaming Luo and Andreas Steiner and Abe Friesen and Abhanshu Sharma and Abheesht Sharma and Adi Mayrav Gilady and Adrian Goedeckemeyer and Alaa Saade and Alex Feng and Alexander Kolesnikov and Alexei Bendebury and Alvin Abdagic and Amit Vadi and András György and André Susano Pinto and Anil Das and Ankur Bapna and Antoine Miech and Antoine Yang and Antonia Paterson and Ashish Shenoy and Ayan Chakrabarti and Bilal Piot and Bo Wu and Bobak Shahriari and Bryce Petrini and Charlie Chen and Charline Le Lan and Christopher A. Choquette-Choo and CJ Carey and Cormac Brick and Daniel Deutsch and Danielle Eisenbud and Dee Cattle and Derek Cheng and Dimitris Paparas and Divyashree Shivakumar Sreepathihalli and Doug Reid and Dustin Tran and Dustin Zelle and Eric Noland and Erwin Huizenga and Eugene Kharitonov and Frederick Liu and Gagik Amirkhanyan and Glenn Cameron and Hadi Hashemi and Hanna Klimczak-Plucińska and Harman Singh and Harsh Mehta and Harshal Tushar Lehri and Hussein Hazimeh and Ian Ballantyne and Idan Szpektor and Ivan Nardini and Jean Pouget-Abadie and Jetha Chan and Joe Stanton and John Wieting and Jonathan Lai and Jordi Orbay and Joseph Fernandez and Josh Newlan and Ju-yeong Ji and Jyotinder Singh and Kat Black and Kathy Yu and Kevin Hui and Kiran Vodrahalli and Klaus Greff and Linhai Qiu and Marcella Valentine and Marina Coelho and Marvin Ritter and Matt Hoffman and Matthew Watson and Mayank Chaturvedi and Michael Moynihan and Min Ma and Nabila Babar and Natasha Noy and Nathan Byrd and Nick Roy and Nikola Momchev and Nilay Chauhan and Noveen Sachdeva and Oskar Bunyan and Pankil Botarda and Paul Caron and Paul Kishan Rubenstein and Phil Culliton and Philipp Schmid and Pier Giuseppe Sessa and Pingmei Xu and Piotr Stanczyk and Pouya Tafti and Rakesh Shivanna and Renjie Wu and Renke Pan and Reza Rokni and Rob Willoughby and Rohith Vallu and Ryan Mullins and Sammy Jerome and Sara Smoot and Sertan Girgin and Shariq Iqbal and Shashir Reddy and Shruti Sheth and Siim Põder and Sijal Bhatnagar and Sindhu Raghuram Panyam and Sivan Eiger and Susan Zhang and Tianqi Liu and Trevor Yacovone and Tyler Liechty and Uday Kalra and Utku Evci and Vedant Misra and Vincent Roseberry and Vlad Feinberg and Vlad Kolesnikov and Woohyun Han and Woosuk Kwon and Xi Chen and Yinlam Chow and Yuvein Zhu and Zichuan Wei and Zoltan Egyed and Victor Cotruta and Minh Giang and Phoebe Kirk and Anand Rao and Kat Black and Nabila Babar and Jessica Lo and Erica Moreira and Luiz Gustavo Martins and Omar Sanseviero and Lucas Gonzalez and Zach Gleicher and Tris Warkentin and Vahab Mirrokni and Evan Senter and Eli Collins and Joelle Barral and Zoubin Ghahramani and Raia Hadsell and Yossi Matias and D. Sculley and Slav Petrov and Noah Fiedel and Noam Shazeer and Oriol Vinyals and Jeff Dean and Demis Hassabis and Koray Kavukcuoglu and Clement Farabet and Elena Buchatskaya and Jean-Baptiste Alayrac and Rohan Anil and Dmitry and Lepikhin and Sebastian Borgeaud and Olivier Bachem and Armand Joulin and Alek Andreev and Cassidy Hardin and Robert Dadashi and Léonard Hussenot},
      year={2025},
      eprint={2503.19786},
      archivePrefix={arXiv},
      primaryClass={cs.CL},
      url={https://arxiv.org/abs/2503.19786}, 
}

@misc{mistralai2025magistral,
      title={Magistral}, 
      author={Mistral-AI and : and Abhinav Rastogi and Albert Q. Jiang and Andy Lo and Gabrielle Berrada and Guillaume Lample and Jason Rute and Joep Barmentlo and Karmesh Yadav and Kartik Khandelwal and Khyathi Raghavi Chandu and Léonard Blier and Lucile Saulnier and Matthieu Dinot and Maxime Darrin and Neha Gupta and Roman Soletskyi and Sagar Vaze and Teven Le Scao and Yihan Wang and Adam Yang and Alexander H. Liu and Alexandre Sablayrolles and Amélie Héliou and Amélie Martin and Andy Ehrenberg and Anmol Agarwal and Antoine Roux and Arthur Darcet and Arthur Mensch and Baptiste Bout and Baptiste Rozière and Baudouin De Monicault and Chris Bamford and Christian Wallenwein and Christophe Renaudin and Clémence Lanfranchi and Darius Dabert and Devon Mizelle and Diego de las Casas and Elliot Chane-Sane and Emilien Fugier and Emma Bou Hanna and Gauthier Delerce and Gauthier Guinet and Georgii Novikov and Guillaume Martin and Himanshu Jaju and Jan Ludziejewski and Jean-Hadrien Chabran and Jean-Malo Delignon and Joachim Studnia and Jonas Amar and Josselin Somerville Roberts and Julien Denize and Karan Saxena and Kush Jain and Lingxiao Zhao and Louis Martin and Luyu Gao and Lélio Renard Lavaud and Marie Pellat and Mathilde Guillaumin and Mathis Felardos and Maximilian Augustin and Mickaël Seznec and Nikhil Raghuraman and Olivier Duchenne and Patricia Wang and Patrick von Platen and Patryk Saffer and Paul Jacob and Paul Wambergue and Paula Kurylowicz and Pavankumar Reddy Muddireddy and Philomène Chagniot and Pierre Stock and Pravesh Agrawal and Romain Sauvestre and Rémi Delacourt and Sanchit Gandhi and Sandeep Subramanian and Shashwat Dalal and Siddharth Gandhi and Soham Ghosh and Srijan Mishra and Sumukh Aithal and Szymon Antoniak and Thibault Schueller and Thibaut Lavril and Thomas Robert and Thomas Wang and Timothée Lacroix and Valeriia Nemychnikova and Victor Paltz and Virgile Richard and Wen-Ding Li and William Marshall and Xuanyu Zhang and Yunhao Tang},
      year={2025},
      eprint={2506.10910},
      archivePrefix={arXiv},
      primaryClass={cs.CL},
      url={https://arxiv.org/abs/2506.10910}, 
}

@misc{microsoft2025phi4minitechnicalreportcompact,
      title={Phi-4-Mini Technical Report: Compact yet Powerful Multimodal Language Models via Mixture-of-LoRAs}, 
      author={Microsoft and : and Abdelrahman Abouelenin and Atabak Ashfaq and Adam Atkinson and Hany Awadalla and Nguyen Bach and Jianmin Bao and Alon Benhaim and Martin Cai and Vishrav Chaudhary and Congcong Chen and Dong Chen and Dongdong Chen and Junkun Chen and Weizhu Chen and Yen-Chun Chen and Yi-ling Chen and Qi Dai and Xiyang Dai and Ruchao Fan and Mei Gao and Min Gao and Amit Garg and Abhishek Goswami and Junheng Hao and Amr Hendy and Yuxuan Hu and Xin Jin and Mahmoud Khademi and Dongwoo Kim and Young Jin Kim and Gina Lee and Jinyu Li and Yunsheng Li and Chen Liang and Xihui Lin and Zeqi Lin and Mengchen Liu and Yang Liu and Gilsinia Lopez and Chong Luo and Piyush Madan and Vadim Mazalov and Arindam Mitra and Ali Mousavi and Anh Nguyen and Jing Pan and Daniel Perez-Becker and Jacob Platin and Thomas Portet and Kai Qiu and Bo Ren and Liliang Ren and Sambuddha Roy and Ning Shang and Yelong Shen and Saksham Singhal and Subhojit Som and Xia Song and Tetyana Sych and Praneetha Vaddamanu and Shuohang Wang and Yiming Wang and Zhenghao Wang and Haibin Wu and Haoran Xu and Weijian Xu and Yifan Yang and Ziyi Yang and Donghan Yu and Ishmam Zabir and Jianwen Zhang and Li Lyna Zhang and Yunan Zhang and Xiren Zhou},
      year={2025},
      eprint={2503.01743},
      archivePrefix={arXiv},
      primaryClass={cs.CL},
      url={https://arxiv.org/abs/2503.01743}, 
}

@misc{miao2025spanbenchmarkingimprovingcrosscalendar,
      title={SPAN: Benchmarking and Improving Cross-Calendar Temporal Reasoning of Large Language Models}, 
      author={Zhongjian Miao and Hao Fu and Chen Wei},
      year={2025},
      eprint={2511.09993},
      archivePrefix={arXiv},
      primaryClass={cs.AI},
      url={https://arxiv.org/abs/2511.09993}, 
}

@inproceedings{suarez2019asynchronous,
  title={Asynchronous pipeline for processing huge corpora on medium to low resource infrastructures},
  author={Su{\'a}rez, Pedro Javier Ortiz and Sagot, Beno{\^\i}t and Romary, Laurent},
  booktitle={7th Workshop on the Challenges in the Management of Large Corpora (CMLC-7)},
  year={2019},
  organization={Leibniz-Institut f{\"u}r Deutsche Sprache}
}

@article{penedo2025fineweb2,
  title={FineWeb2: One Pipeline to Scale Them All--Adapting Pre-Training Data Processing to Every Language},
  author={Penedo, Guilherme and Kydl{\'\i}{\v{c}}ek, Hynek and Sabol{\v{c}}ec, Vinko and Messmer, Bettina and Foroutan, Negar and Kargaran, Amir Hossein and Raffel, Colin and Jaggi, Martin and Von Werra, Leandro and Wolf, Thomas},
  journal={arXiv preprint arXiv:2506.20920},
  year={2025}
}

@software{alshehri_2024_11288565,
  author       = {Alshehri, Mohammed H},
  title        = {HijriDate: A Python package for Hijri-Gregorian
                   date conversion
                  },
  month        = may,
  year         = 2024,
  publisher    = {Zenodo},
  version      = {2.5.0},
  doi          = {10.5281/zenodo.11288565},
  url          = {https://doi.org/10.5281/zenodo.11288565},
}

@inproceedings{mazzia-etal-2026-benchmarking,
    title = "Benchmarking Multilingual Temporal Reasoning in {LLM}s: The Temporal Reasoning Dataset",
    author = "Mazzia, Vittorio  and
      Pollastrini, Sandro  and
      Bernardi, Davide  and
      Rubagotti, Chiara  and
      Amberti, Daniele",
    editor = "Riccardi, Giuseppe  and
      Mousavi, Seyed Mahed  and
      Torres, Maria Ines  and
      Yoshino, Koichiro  and
      Callejas, Zoraida  and
      Chowdhury, Shammur Absar  and
      Chen, Yun-Nung  and
      Bechet, Frederic  and
      Gustafson, Joakim  and
      Damnati, G{\'e}raldine  and
      Papangelis, Alex  and
      D{'}Haro, Luis Fernando  and
      Mendon{\c{c}}a, John  and
      Bernardi, Raffaella  and
      Hakkani-Tur, Dilek  and
      Di Fabbrizio, Giuseppe {''}Pino{''}  and
      Kawahara, Tatsuya  and
      Alam, Firoj  and
      Tur, Gokhan  and
      Johnston, Michael",
    booktitle = "Proceedings of the 16th International Workshop on Spoken Dialogue System Technology",
    month = feb,
    year = "2026",
    address = "Trento, Italy",
    publisher = "Association for Computational Linguistics",
    url = "https://aclanthology.org/2026.iwsds-1.19/",
    pages = "168--181"
}

@inproceedings{bhatia-etal-2025-date,
    title = "Date Fragments: A Hidden Bottleneck of Tokenization for Temporal Reasoning",
    author = "Bhatia, Gagan  and
      Peyrard, Maxime  and
      Zhao, Wei",
    editor = "Christodoulopoulos, Christos  and
      Chakraborty, Tanmoy  and
      Rose, Carolyn  and
      Peng, Violet",
    booktitle = "Proceedings of the 2025 Conference on Empirical Methods in Natural Language Processing",
    month = nov,
    year = "2025",
    address = "Suzhou, China",
    publisher = "Association for Computational Linguistics",
    url = "https://aclanthology.org/2025.emnlp-main.159/",
    doi = "10.18653/v1/2025.emnlp-main.159",
    pages = "3201--3219",
    ISBN = "979-8-89176-332-6"
}

@inproceedings{bhatia-etal-2025-datelogicqa,
    title = "{D}ate{L}ogic{QA}: Benchmarking Temporal Biases in Large Language Models",
    author = "Bhatia, Gagan  and
      Tang, Ming Ze  and
      Mahanta, Cristina  and
      Kazi, Madiha and 
      Peyrard, Maxime and 
      Zhao, Wei",
    editor = "Ebrahimi, Abteen  and
      Haider, Samar  and
      Liu, Emmy  and
      Haider, Sammar  and
      Leonor Pacheco, Maria  and
      Wein, Shira",
    booktitle = "Proceedings of the 2025 Conference of the Nations of the Americas Chapter of the Association for Computational Linguistics: Human Language Technologies (Volume 4: Student Research Workshop)",
    month = apr,
    year = "2025",
    address = "Albuquerque, USA",
    publisher = "Association for Computational Linguistics",
    url = "https://aclanthology.org/2025.naacl-srw.32/",
    doi = "10.18653/v1/2025.naacl-srw.32",
    pages = "321--332",
    ISBN = "979-8-89176-192-6"
}

@misc{liu2025temporaltokenizationstrategiesevent,
      title={Temporal Tokenization Strategies for Event Sequence Modeling with Large Language Models}, 
      author={Zefang Liu and Nam H. Nguyen and Yinzhu Quan and Shi-Xiong Zhang},
      year={2025},
      eprint={2512.13618},
      archivePrefix={arXiv},
      primaryClass={cs.CL},
      url={https://arxiv.org/abs/2512.13618}, 
}

@inproceedings{sasaki-etal-2025-language,
    title = "Can Language Models Handle a Non-Gregorian Calendar? The Case of the {J}apanese wareki",
    author = "Sasaki, Mutsumi  and
      Kamoda, Go  and
      Takahashi, Ryosuke  and
      Sato, Kosuke  and
      Inui, Kentaro  and
      Sakaguchi, Keisuke  and
      Heinzerling, Benjamin",
    editor = "Inui, Kentaro  and
      Sakti, Sakriani  and
      Wang, Haofen  and
      Wong, Derek F.  and
      Bhattacharyya, Pushpak  and
      Banerjee, Biplab  and
      Ekbal, Asif  and
      Chakraborty, Tanmoy  and
      Singh, Dhirendra Pratap",
    booktitle = "Proceedings of the 14th International Joint Conference on Natural Language Processing and the 4th Conference of the Asia-Pacific Chapter of the Association for Computational Linguistics",
    month = dec,
    year = "2025",
    address = "Mumbai, India",
    publisher = "The Asian Federation of Natural Language Processing and The Association for Computational Linguistics",
    url = "https://aclanthology.org/2025.ijcnlp-short.36/",
    pages = "444--463",
    ISBN = "979-8-89176-299-2"
}

@misc{pezik2025llmlagbenchidentifyingtemporaltraining,
      title={LLMLagBench: Identifying Temporal Training Boundaries in Large Language Models}, 
      author={Piotr Pęzik and Konrad Kaczyński and Maria Szymańska and Filip Żarnecki and Zuzanna Deckert and Jakub Kwiatkowski and Wojciech Janowski},
      year={2025},
      eprint={2511.12116},
      archivePrefix={arXiv},
      primaryClass={cs.CL},
      url={https://arxiv.org/abs/2511.12116}, 
}

@inproceedings{holtermann-etal-2025-around,
    title = "Around the World in 24 Hours: Probing {LLM} Knowledge of Time and Place",
    author = {Holtermann, Carolin  and
      R{\"o}ttger, Paul  and
      Lauscher, Anne},
    editor = "Che, Wanxiang  and
      Nabende, Joyce  and
      Shutova, Ekaterina  and
      Pilehvar, Mohammad Taher",
    booktitle = "Proceedings of the 63rd Annual Meeting of the Association for Computational Linguistics (Volume 1: Long Papers)",
    month = jul,
    year = "2025",
    address = "Vienna, Austria",
    publisher = "Association for Computational Linguistics",
    url = "https://aclanthology.org/2025.acl-long.1115/",
    doi = "10.18653/v1/2025.acl-long.1115",
    pages = "22875--22897",
    ISBN = "979-8-89176-251-0"
}

@misc{wang2026measuringiterativetemporalreasoning,
      title={Measuring Iterative Temporal Reasoning with Time Puzzles}, 
      author={Zhengxiang Wang and Zeyu Dong},
      year={2026},
      eprint={2601.07148},
      archivePrefix={arXiv},
      primaryClass={cs.CL},
      url={https://arxiv.org/abs/2601.07148}, 
}

\appendix

\section{Appendix}
\label{sec:appendix}

\subsection{Creation of our \dataset}
\label{app:date_pipeline}

To ensure consistency across the multilingual benchmark, we implemented a unified processing pipeline. This pipeline processes the English source data and generates language-specific variants for Arabic, Chinese, Hausa, German, and English. The process consists of two stages: \textit{Standardization} and \textit{Polymorphic Formatting}.

\subsubsection{Stage 1: Date Extraction and Standardization}
The first step is identical for all five languages. We utilize a regular expression to identify date entities within the source text. Regardless of the input format, these dates are parsed into a standard internal representation (Year, Month, Day). This ensures that all downstream formatters operate on a consistent temporal grounding.

\subsubsection{Stage 2: Polymorphic Formatting}
Once standardized, the pipeline applies four distinct formatters per language: \textbf{ISO}, \textbf{Slash} (Numeric), \textbf{Long} (Textual), and \textbf{Calendar} (Phrasal/Cultural). The specific logic for each language is detailed below. The conversion process was implemented through a unified Python pipeline. For each language, the system first extracts and parses dates from the source English questions into a standard internal representation. Language-specific formatters are then applied. For instance, Arabic formatting involves converting digits to Arabic-Indic numerals, applying right-to-left marks for ISO dates, and using the hijri-converter library to generate Hijri calendar dates (e.g., \begin{RLtext}"ذو الحجة ١٤٤٤هـ"\end{RLtext}). Similarly, Chinese formatting integrates conversions to the traditional lunar calendar.

\paragraph{Arabic Implementation.}
The Arabic formatting pipeline requires specific handling for text directionality and numeral systems.
\begin{itemize}
    \item \textbf{ISO Format:} To prevent rendering issues in Right-to-Left (RTL) contexts, the standard ISO string is wrapped in Unicode Left-to-Right Marks (LRM, \texttt{U+200E}).
    \item \textbf{Long Format:} We map Gregorian month indices to their Arabic counterparts (e.g., \textit{July} $\rightarrow$ \textit{Yuliyu}) and convert Western Arabic numerals (0-9) to Eastern Arabic-Indic numerals (\RL{٠-٩}).
    \item \textbf{Calendar (Hijri) Format:} We utilize the \texttt{hijri-converter} library to transform the Gregorian date into the Hijri calendar. The resulting day, month, and year are formatted using standard Hijri month names (e.g., \textit{Ramadan}, \textit{Shawwal}).
\end{itemize}

\paragraph{Chinese Implementation.}
Chinese formatting emphasizes the use of component suffixes and Lunar conversion.
\begin{itemize}
    \item \textbf{Long Format:} Adheres to the standard East Asian order (Year-Month-Day) with the respective character suffixes (\begin{CJK*}{UTF8}{gbsn}年, 月, 日\end{CJK*}).
    \item \textbf{Calendar (Lunar) Format:} We convert the Gregorian date to the Chinese Lunar calendar using the \texttt{lunarcalendar} library. The numeric years are converted to their Chinese character equivalents (e.g., 2023 $\rightarrow$ \begin{CJK*}{UTF8}{gbsn}二零二三\end{CJK*}), and months are mapped to their traditional lunar representations.
\end{itemize}

\paragraph{Hausa Implementation.}
Hausa formatting integrates Islamic cultural elements with standard Gregorian tracking, reflecting the region's dual-calendar usage.
\begin{itemize}
    \item \textbf{Long Format:} Uses the particle ``ga'' (meaning ``on'') to connect the day and the month (e.g., \textit{03 ga Afrilu 2023}).
    \item \textbf{Calendar Format:} In this variant, we utilize the locally recognized Islamic month names (e.g., \textit{Ramadan}, \textit{Shawwal}) while maintaining the Gregorian year for clarity in civil contexts.
\end{itemize}

\paragraph{German Implementation.}
German requires specific grammatical phrasings for the ``Calendar'' variant to represent a formal date expression. While the standard formats use dot separators (DD.MM.YYYY), the calendar variant expands this to a formal phrase: ``Am [Day]. [Month] des Jahres [Year]'' (e.g., \textit{Am 26. Juni des Jahres 2025}).

\subsubsection{Examples}

\begin{table*}[ht]
\centering
\footnotesize
\setlength\tabcolsep{3pt} 
\begin{tabularx}{\textwidth}{@{} l c c c c X l @{}}
\toprule
\textbf{Task} & \textbf{Raw} & \textbf{Fmt} & \textbf{Lng} & \textbf{Size} & \textbf{Example} & \textbf{GT} \\
\midrule
Arithmetic & 250 & 4 & 5 & 5,000 & In a movie, the tower took exactly 14 years to construct. They started in 2000-12-27. When was it ready? & 2014-12-27 \\
Time Zone & 250 & 4 & 5 & 5,000 & If it’s 2 AM on 1352-03-02 in Asia/Singapore, what’s the date and time in Europe/Athens? &  8 PM on 1352-03-01 \\
Relation & 250 & 4 & 5 & 5,000 & Rules for lending against stocks and unit trusts were also redefined. What is the relationship between the event `redefined' and the time `April 1, 1997'?  & IS\_INCLUDED \\
\midrule
\textbf{Total} & \textbf{750} & \textbf{6} & \textbf{5} & \textbf{15,000} & & \\
\bottomrule
\end{tabularx}
\caption{\textbf{Overview of tasks in the \dataset dataset.} ``Raw'' denotes unique English questions. ``Size'' is the total number of examples after multilingual/format expansion ($Raw \times 4 \text{ Fmt} \times 5 \text{ Lang}$). The Truth column shows the expected answer format. We have $6$ unique date formats. }
\label{tab:tasks}
\end{table*}

Example of our \dataset is provided in Table~\ref{tab:tasks}. 

\subsection{Validation of Multilingual Date Fragmentation Ratio (mDFR)}
\label{app:metric_validation}

This appendix provides a detailed account of the formulation and two-part validation process for our custom Multilingual Date Fragmentation Ratio (mDFR). We demonstrate that this metric aligns closely with human intuition regarding semantic disruption and relies on empirically sound weightings. 

\subsubsection{Metric Formulation}
We calculate the structural divergence $\theta$ between the model's token count vector $\mathbf{t}$ and the semantic baseline vector $\mathbf{b}$ using cosine distance. This metric quantifies the deviation of the model's tokenisation from an ideal semantic segmentation. The final mDFR score, $F \in [0,1]$, is constructed as a weighted sum of four specific error components: whether semantic roots are split ($\mathbbm{1}_{\mathrm{split}}$), whether delimiters are lost ($\mathbbm{1}_{\mathrm{delimiter}}$), the increase in total token count ($\Delta N$), and the distributional divergence ($\theta$).

\subsubsection{Human Evaluation of Fragmentation Severity}
\label{app:human_eval}
This study was designed to confirm that our F metric captures what humans perceive as semantic disruption in tokenized dates more effectively than general-purpose text similarity metrics.

\paragraph{Methodology.} We recruited five computer science graduate students, who were familiar with NLP but blind to our hypotheses, to serve as annotators. We created a stimulus set of 100 tokenised date strings, stratified to represent a wide range of models, date formats, and fragmentation levels from our experiments. For each item, annotators were shown the original date and the list of sub-tokens, and asked to rate the \textbf{“fragmentation severity”} on a 5-point Likert scale, according to the following rubric:
\begin{itemize}
    \item \textbf{1 (No Fragmentation):} Tokens perfectly preserve the semantic components.
    \item \textbf{2 (Minor Fragmentation):} Mostly preserved, with minor, non-ideal splits.
    \item \textbf{3 (Moderate Fragmentation):} Core components are broken, making the structure harder to discern. Delimiters might be lost or numbers oddly grouped.
    \item \textbf{4 (High Fragmentation):} Date split into many small pieces (e.g., single digits), though the original characters are easily reassembled.
    \item \textbf{5 (Severe Fragmentation):} tokenisation completely obscures the date's structure, often by adding non-numeric tokens or creating highly unintuitive groupings.
\end{itemize}
The human judgments were highly reliable, with a Krippendorff's Alpha for inter-annotator agreement of $\alpha = 0.81$.

\paragraph{Results.} We computed the Spearman's rank correlation coefficient ($\rho$) between the average human rating for each item and the scores from our F metric, BLEU, and character-level Edit Distance. As shown in Table \ref{tab:human_eval_appendix}, our F metric demonstrated a strong correlation with human ratings ($\rho=0.89$), far exceeding general-purpose metrics like BLEU ($\rho=0.43$).

\begin{table}[h]
    \centering
    \begin{tabular}{lc}
        \toprule
        \textbf{Metric} & \textbf{Correlation ($\rho$)} \\
        \midrule
        \textbf{mDFR} & \textbf{0.89} \\
        DFR \cite{bhatia-etal-2025-date} & \underline{0.81} \\
        BLEU Score                            & 0.43          \\
        Character-Level Edit Distance         & 0.29          \\
        \bottomrule
    \end{tabular}
    \caption{Spearman Correlation ($\rho$) of Metrics with Human Judgments of Fragmentation Severity for Multilingual dates.}
    \label{tab:human_eval_appendix}
\end{table}

\subsubsection{Data-Driven Validation of Metric Coefficients}
\label{app:data_driven_validation}
To directly tune our metric to align with human perception, we framed the weight determination as a linear regression problem. The goal was to predict the average human severity rating using the four fragmentation components as features: $\mathbf{x} = [\mathbbm{1}_{\text{split}}, \mathbbm{1}_{\text{delimiter}}, (N - N_b), \theta]$.

After fitting the model to our human evaluation data, we obtained a set of empirically derived coefficients. As shown in Table \ref{tab:learned_weights_appendix}, the weights learned from human ratings are remarkably similar to the normalised version of our original, intuitively set weights. This confirms that Distributional Divergence ($\theta$) is the dominant factor in perceived severity, followed by structural breaks, with token count inflation playing a minor role.

\begin{table*}[t]
    \centering
    \footnotesize
    \begin{tabular}{lcc}
        \toprule
        \textbf{Fragmentation Component} & \textbf{Original Intuitive Weight} & \textbf{Empirically Learned Weight} \\
        & \textbf{(Normalised)} & \textbf{(from Human Ratings)} \\
        \midrule
        $1_{\text{split}}$ (Component Split)    & $0.1818$ & \textbf{0.2015} \\
        $1_{\text{delimiter}}$ (Delimiter Loss) & $0.1818$ & \textbf{0.1932} \\
        $N - N_b$ (Token Difference)            & $0.0909$ & \textbf{0.1053} \\
        $\theta$ (Distributional Divergence)    & $0.5455$ & \textbf{0.5000} \\
        \bottomrule
    \end{tabular}
    \caption{Comparison of Original (Normalised) and Empirically Learned Weights for the F Metric.}
    \label{tab:learned_weights_appendix}
\end{table*}

\subsubsection{Qualitative Analysis of Fragmentation}
\label{app:qualitative_analysis}
To visualise how mDFR scores correspond to real-world model outputs, we analysed tokenisation patterns across different languages and scripts. Table \ref{tab:token_qualitative} illustrates the correlation between high mDFR scores, human severity ratings, and severe segmentation issues. Notably, non-Latin scripts (e.g., Arabic, Chinese) and agglutinative languages often suffer from higher fragmentation (rated 4.6–5.0 by humans), where semantic roots are often shattered into single characters or bytes.

\begin{table*}[!htp]
\centering
\resizebox{\textwidth}{!}{
\begin{tabular}{llclllcc}
\toprule
\textbf{Format} & \textbf{Language} & \textbf{Calendar} & \textbf{Original String} & \textbf{Baseline tokenisation} & \textbf{Gemma 3 tokenisation (Visualized)} & \textbf{mDFR} & \textbf{Avg. Human Rating} \\
\midrule
DD. Month YYYY & German & Greg. & 10. Oktober 2034 & 10 . Oktober 2034 & \texttt{1} $|$ \texttt{0} $|$ \texttt{.} $|$ \texttt{Oktober} $|$ \texttt{2} $|$ \texttt{0} $|$ \texttt{3} $|$ \texttt{4} & 0.50 & 4.2 \\
Month DD, YYYY & English & Greg. & October 10, 2034 & October 10 , 2034 & \texttt{October} $|$ \texttt{1} $|$ \texttt{0} $|$ \texttt{,} $|$ \texttt{2} $|$ \texttt{0} $|$ \texttt{3} $|$ \texttt{4} & 0.53 & 4.4 \\
\begin{CJK*}{UTF8}{gbsn}YYYY年MM月DD日\end{CJK*} & Chinese & Greg. & \begin{CJK*}{UTF8}{gbsn}2034 年 10 月 10 日\end{CJK*} & \begin{CJK*}{UTF8}{gbsn}2034 年 10 月 10 日\end{CJK*} & \texttt{2} $|$ \texttt{0} $|$ \texttt{3} $|$ \texttt{4} $|$ \begin{CJK*}{UTF8}{gbsn}年\end{CJK*} $|$ \texttt{1} $|$ \texttt{0} $|$ \begin{CJK*}{UTF8}{gbsn}月\end{CJK*} $|$ \texttt{1} $|$ \texttt{0} $|$ \begin{CJK*}{UTF8}{gbsn}日\end{CJK*} & 0.55 & 4.6 \\
DD Month YYYY \RL{هـ} & Arabic & Hijri & \RL{٢٧ رجب ١٤٥٦ هـ} & \RL{٢٧ رجب ١٤٥٦ هـ} & \RL{٢} $|$ \RL{٧} $|$ \RL{ر} $|$ \RL{جب} $|$ \RL{١} $|$ \RL{٤} $|$ \RL{٥} $|$ \RL{٦} $|$ \RL{هـ} & 0.60 & 5.0 \\
DD Month YYYY AH & English & Hijri & 27 Rajab 1456 AH & 27 Rajab 1456 AH & \texttt{2} $|$ \texttt{7} $|$ \texttt{Raj} $|$ \texttt{ab} $|$ \texttt{1} $|$ \texttt{4} $|$ \texttt{5} $|$ \texttt{6} $|$ \texttt{AH} & 0.60 & 5.0 \\
DD Month YYYY & Arabic & Greg. & \RL{١٠ أكتوبر ٢٠٣٤} & \RL{١٠ أكتوبر ٢٠٣٤} & \RL{١} $|$ \RL{٠} $|$ \RL{أكتوبر} $|$ \RL{٢} $|$ \RL{٠} $|$ \RL{٣} $|$ \RL{٤} & 0.70 & 5.0 \\
DD Month YYYY & English & Greg. & 10 October 2034 & 10 October 2034 & \texttt{1} $|$ \texttt{0} $|$ \texttt{October} $|$ \texttt{2} $|$ \texttt{0} $|$ \texttt{3} $|$ \texttt{4} & 0.75 & 5.0 \\
Month DD, YYYY & Hausa & Greg. & Oktoba 10, 2034 & Oktoba 10 , 2034 & \texttt{O} $|$ \texttt{kt} $|$ \texttt{oba} $|$ \texttt{1} $|$ \texttt{0} $|$ \texttt{,} $|$ \texttt{2} $|$ \texttt{0} $|$ \texttt{3} $|$ \texttt{4} & 0.78 & 5.0 \\
\bottomrule
\end{tabular}
}
\caption{\textbf{Qualitative Analysis of Tokenisation Fragmentation.} Vertical bars ($|$) denote token boundaries within the Gemma 3 tokeniser. The \textbf{Avg. Human Rating} (1-5 scale) confirms that higher mDFR scores correspond to perceived severe fragmentation.}
\label{tab:token_qualitative_hr}
\end{table*}

\subsection{Correlation of the different tasks}
\label{app:dfr_other_tasks}

\begin{figure*}[t!]
    \centering
    \includegraphics[width=\textwidth]{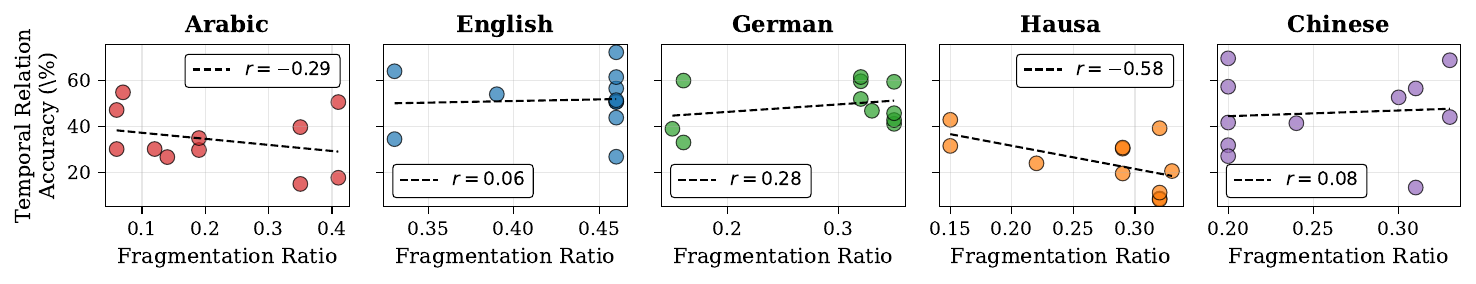}
    \caption{\textbf{Impact of Tokenisation on Temporal Relation Task Accuracy.} The scatter plots show the correlation between Date Fragmentation Ratio (DFR) and temporal reasoning accuracy for each language.}
    \label{fig:dfr_correlation_tr}
\end{figure*}

\begin{figure*}[t!]
    \centering
    \includegraphics[width=\textwidth]{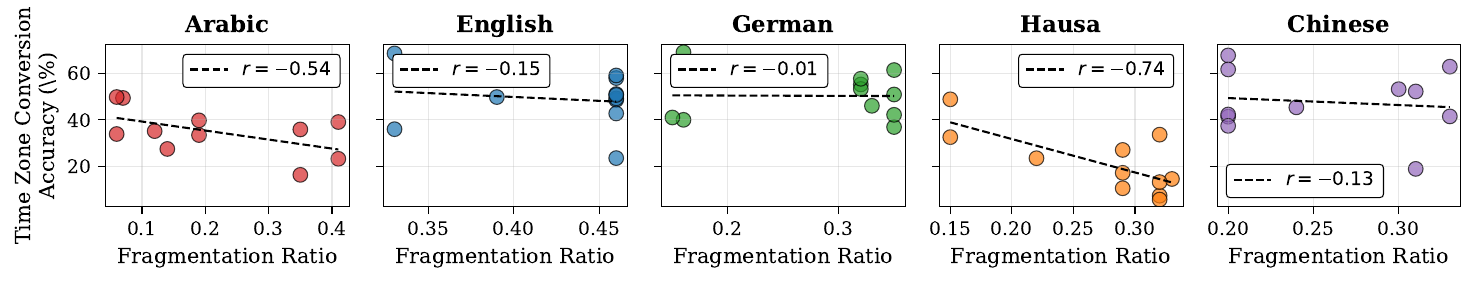}
    \caption{\textbf{Impact of Tokenisation on Time Zone Conversion Task Accuracy.} The scatter plots show the correlation between Date Fragmentation Ratio (DFR) and temporal reasoning accuracy for each language.}
    \label{fig:dfr_correlation_tz}
\end{figure*}

The same broad pattern holds in the other two tasks, though the strength of the effect varies by task. In \textbf{temporal relation extraction} (Figure~\ref{fig:dfr_correlation_tr}), higher fragmentation is associated with lower accuracy in the two low-resource languages, especially Hausa ($r=-0.58$), while the relationship remains weak or near-zero in English ($r=0.06$), German ($r=0.28$), and Chinese ($r=0.08$). Arabic also shows a modest negative correlation ($r=-0.29$). As in date arithmetic, these results suggest that \textbf{date fragmentation} is more consequential in low-resource settings, whereas high-resource languages are generally more robust to fragmented temporal inputs.

A similar but slightly stronger pattern appears in \textbf{time zone conversion} (Figure~\ref{fig:dfr_correlation_tz}). Fragmentation is negatively correlated with accuracy in Arabic ($r=-0.54$) and especially Hausa ($r=-0.74$), but remains weak in English ($r=-0.15$), German ($r=-0.01$), and Chinese ($r=-0.13$). Compared with temporal relation extraction, time zone conversion shows a clearer low-resource penalty, though still less extreme than the effect observed for date arithmetic. Overall, across all three tasks, \textbf{date fragmentation} is most predictive of failure in low-resource languages, supporting the view that tokenisation is a regime-dependent bottleneck rather than a universal explanation of temporal reasoning errors.

\subsection{Human Evaluation Details.}
\label{app:human_eval}
To validate the reliability of the LLM-based judging pipeline, we conducted a human evaluation on a subset of the benchmark. Six annotators participated in the study, all of whom were Master's students in computer science or closely related disciplines. For each language included in the validation set, at least two annotators independently reviewed the model outputs and determined whether the response should be classified as \texttt{CORRECT}, \texttt{INCORRECT}, or \texttt{NOT\_ATTEMPTED}. The evaluation covered multiple languages present in the benchmark to ensure that linguistic diversity did not bias the assessment. Disagreements were resolved using majority voting across annotators. Across the evaluated instances, the human annotators achieved an average agreement rate of approximately 89\%, indicating strong consistency in the evaluation criteria. This agreement level provides additional confidence in the reliability of the automated LLM-as-a-judge evaluation protocol used in our experiments.

\subsection{Temporal Geometry}
\label{app:internal_repr}

\begin{figure*}[p]
    \centering
    \begin{subfigure}[b]{0.48\textwidth}
        \centering
        \includegraphics[width=\linewidth]{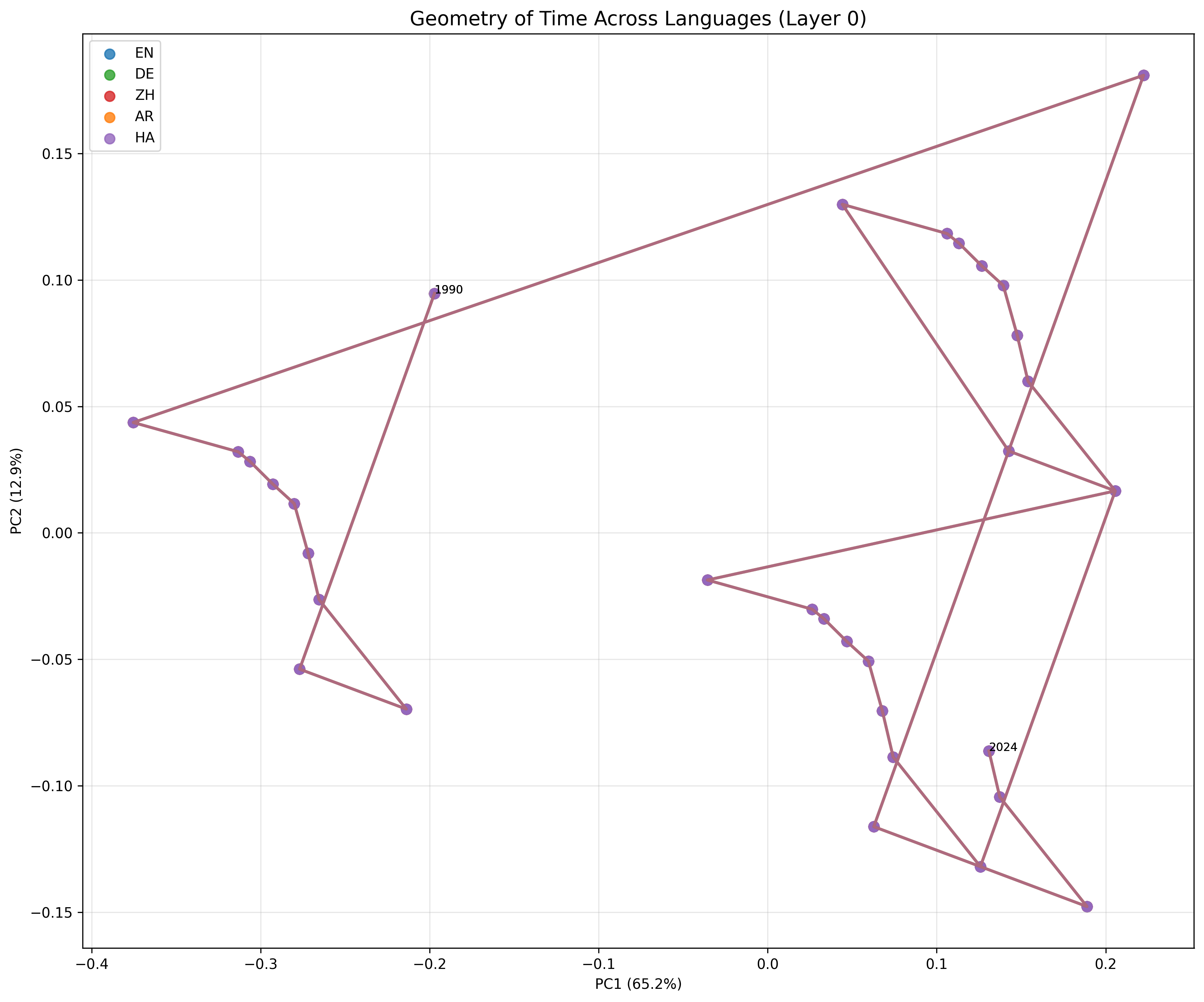}
        \caption{\textbf{Layer 0 (Input):} Chaotic trajectories dominated by surface-level tokenisation fragmentation.}
        \label{fig:pca_layer_0}
    \end{subfigure}
    \hfill
    \begin{subfigure}[b]{0.48\textwidth}
        \centering
        \includegraphics[width=\linewidth]{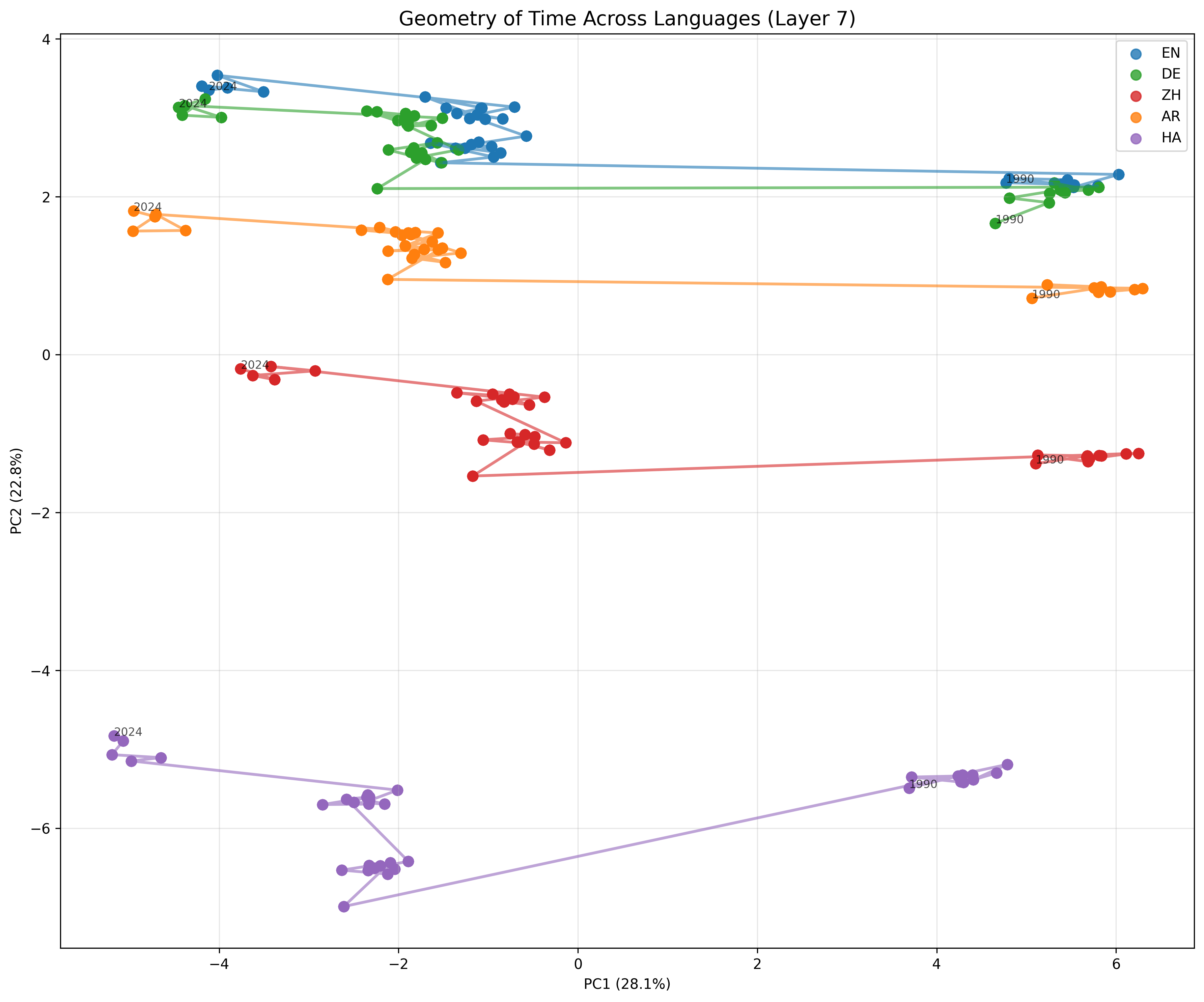}
        \caption{\textbf{Layer 7 (Early):} Representations separate by language syntax; global linear organization has not yet formed.}
        \label{fig:pca_layer_7}
    \end{subfigure}

    \vspace{0.5cm}

    \begin{subfigure}[b]{0.48\textwidth}
        \centering
        \includegraphics[width=\linewidth]{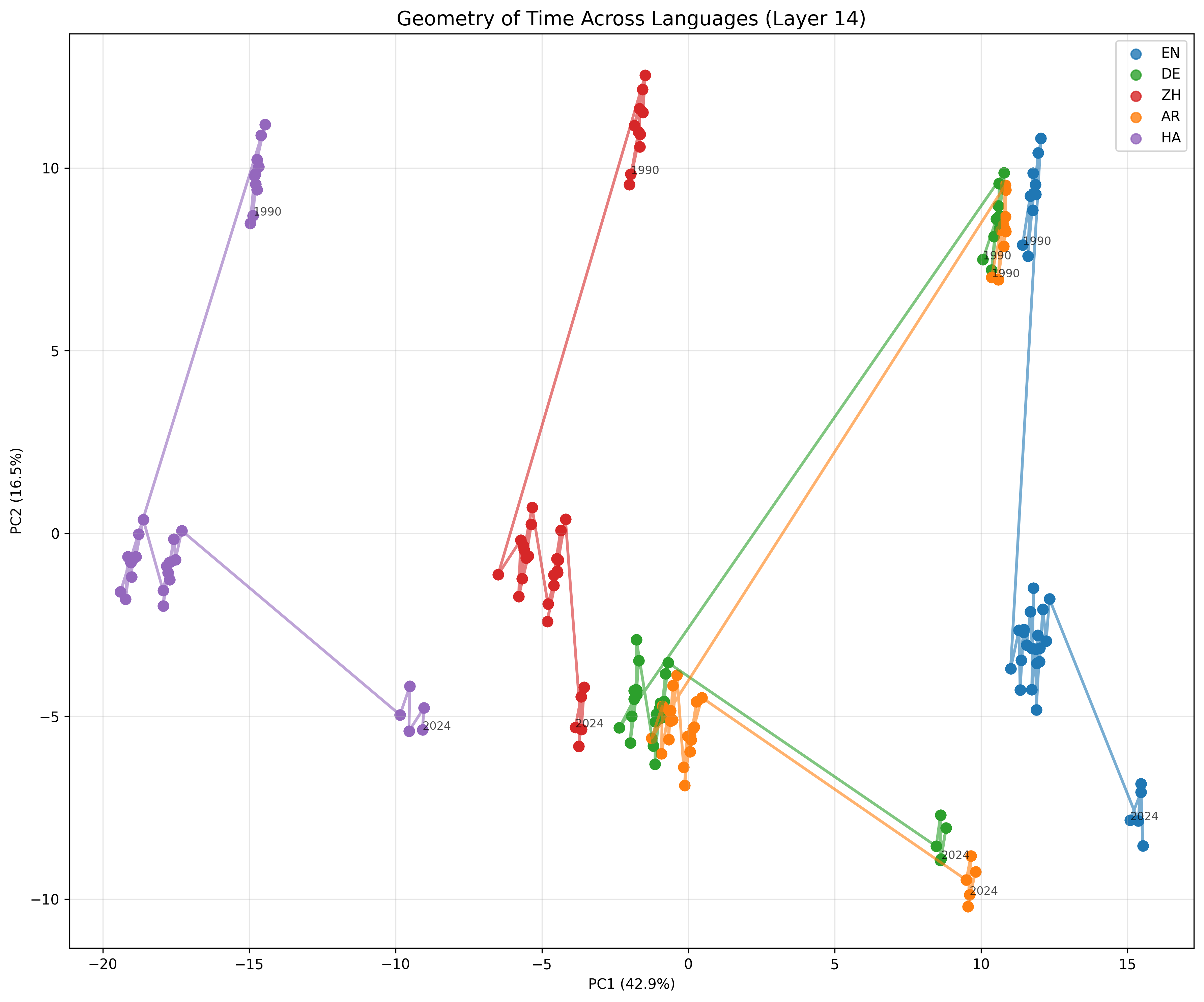}
        \caption{\textbf{Layer 14 (Middle):} High-resource languages (EN, DE, ZH) begin to straighten; low-resource (HA) remains curved.}
        \label{fig:pca_layer_14}
    \end{subfigure}
    \hfill
    \begin{subfigure}[b]{0.48\textwidth}
        \centering
        \includegraphics[width=\linewidth]{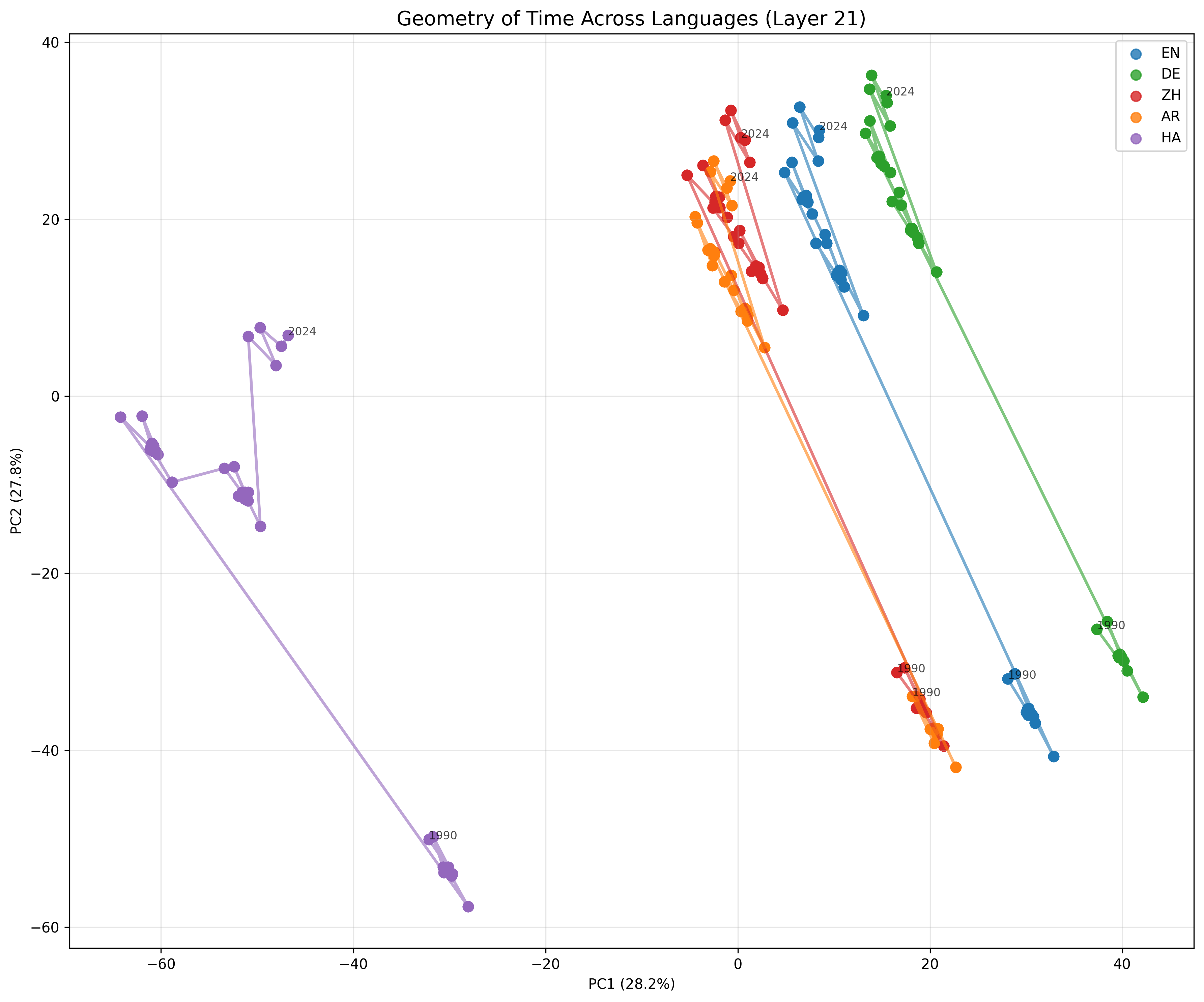}
        \caption{\textbf{Layer 21 (Reasoning):} The \textit{Geometric Language Tax}. EN/DE/ZH form near-linear year trajectories useful for arithmetic, while HA remains a non-linear cluster.}
        \label{fig:pca_layer_21}
    \end{subfigure}

    \vspace{0.5cm}

    \begin{subfigure}[b]{0.48\textwidth}
        \centering
        \includegraphics[width=\linewidth]{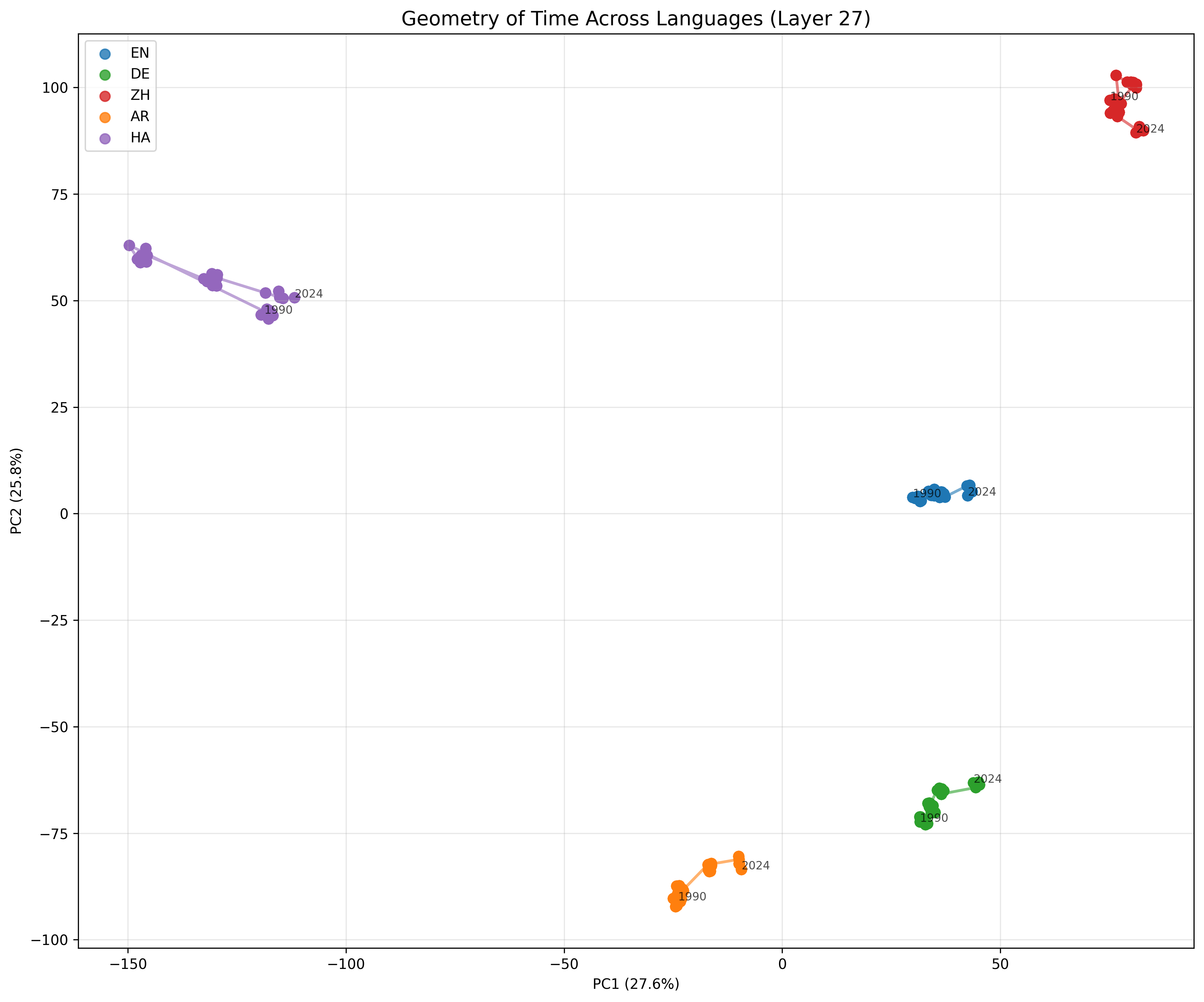}
        \caption{\textbf{Layer 27 (Output):} Final separation of language clusters to prepare for distinct lexical decoding.}
        \label{fig:pca_layer_27}
    \end{subfigure}

    \caption{\textbf{Evolution of temporal organization across layers.}
    PCA projections of year centroid embeddings (1990--2024) in Qwen~3. The plots show a progression from input-level fragmentation (Layer~0) to temporally structured, approximately linear trajectories in mid-to-deep layers for high-resource languages, while Hausa fails to linearize, remaining geometrically misaligned with the year-structured axis.}
    \label{fig:pca_evolution}
\end{figure*}

\paragraph{PCA visualization across layers.}
To provide a qualitative view of how temporal structure emerges across depth, we apply PCA to the set of points $\{\bar{\mathbf{h}}^{(\ell)}_{y,i}\}$ for $y\in[1990,2024]$ across the five languages. In the visualizations (Figure~\ref{fig:pca_evolution}), the resulting plots display the sequence of line segments connecting consecutive years, revealing whether languages form coherent, linear paths in the embedding space.

\subsection{LLM as judge Prompts}
Our prompt for LLM-as-judge is illustrated in Table \ref{tab:eval_prompt}.
\begin{table*}[ht]
    \centering
    \footnotesize
    \begin{tcolorbox}[
        colback=green!5!white,
        colframe=green!75!black,
        title=Human Evaluation]

        \begin{center}
            \Large \textbf{Context-based resolution}
        \end{center}
        \vspace{0.5em}

        \textbf{Prompt}: Who was the chair of Allgemeiner Deutscher Fahrrad-Club in 17/10/2016? \\ 
        \textbf{Gold Answer}: Ulrich Syberg \\ 
        \textbf{Model Prediction}: As of October 17, 2016, the Federal Chairman was Ulrich Syberg\\ 

        \textbf{Human Annotator Rating}: \; \CircledA{CORRECT} \hfill 
        \\[0.2em]
        \textbf{LLM-as-Judge Rating}: \; \CircledA{CORRECT} \hfill 

        \tcblower
        \begin{center}
            \Large \textbf{Date arithmetic}
        \end{center}
        \vspace{0.5em}

        \textbf{Prompt}: What date is 60 days after 05/01/1225? \\
        \textbf{Gold Answer}: March 6, 1225 , June 29, 1225\\
        \textbf{Model Prediction}: July 30, 1225\\

        \textbf{Human Annotator Rating}: \; \CircledC{INCORRECT} \\
        \textbf{LLM-as-Judge Rating}: \; \CircledC{INCORRECT} \\
    \end{tcolorbox}
    \caption{Human evaluation of LLM-as-judge.}
    \label{tab:human_eval}
\end{table*}

\begin{table*}[ht]
  \centering
  \begin{tcolorbox}[
      colback=blue!5!white,
      colframe=blue!75!black,
      title=LLM‑as‑Judge Evaluation Prompt]
\small
\textbf{Your task:} Evaluate one prediction at a time. You receive:
\begin{itemize}
  \item \textbf{Question} – the task prompt shown to the model
  \item \textbf{Gold target} – \emph{all} answers that are considered correct
  \item \textbf{Predicted answer} – the model’s response
\end{itemize}

Return \textbf{one letter only}:\\
\begin{tabular}{lll}
\textbf{A} & CORRECT        & prediction fully matches \emph{one} gold variant          \\
\textbf{B} & INCORRECT      & prediction contradicts or misses required info             \\
\textbf{C} & NOT\_ATTEMPTED & prediction refuses, guesses, or answers irrelevantly        \\
\end{tabular}
\\
[1.5pt]

\textbf{General rules}:
\begin{enumerate}\itemsep2pt
  \item Match semantics, ignore capitalisation, punctuation, order.
  \item If any statement contradicts the gold target, grade \textbf{B}.
  \item Hedging ("I think…") is fine if the correct info is present and no incorrect info is added.
  \item Partial answers are \textbf{B}. Typos that preserve meaning are allowed.
\end{enumerate}
\textbf{DateAugBench specifics}:\\[-6pt]
\begin{itemize}\itemsep2pt
  \item \textbf{Date format ambiguity}: gold lists every valid interpretation; accept any.
  \item \textbf{Date arithmetic}: prediction must match \emph{day, month, year} of a listed variant, any textual format allowed.
  \item \textbf{Format‑switch questions}: answer with any synonym of \texttt{Yes/True} or \texttt{No/False}.
  \item \textbf{Numeric answers} – must match the gold number to the last shown significant digit.
\end{itemize}

\textbf{Output format}\\
Return exactly one capital letter:
\[
\texttt{A}\quad\text{or}\quad\texttt{B}\quad\text{or}\quad\texttt{C}
\]
No additional text or punctuation.

\medskip
\textbf{Example template}
\begin{verbatim}
Question: {question}
Gold target: {target}
Predicted answer: {predicted_answer}
\end{verbatim}

\medskip
\textbf{Now grade:}
\[
\texttt{A}\quad\text{or}\quad\texttt{B}\quad\text{or}\quad\texttt{C}
\]
  \end{tcolorbox}
  \caption{LLM-as-Judge prompt used for comparing model and gold answers in the three tasks in \dataset.}
  \label{tab:eval_prompt}
\end{table*}

\end{document}